\documentclass[preprint,journal]{vgtc}       % preprint (journal style)

\ifpdf%                                % if we use pdflatex
  \pdfoutput=1\relax                   % create PDFs from pdfLaTeX
  \usepackage{graphicx}                % allow us to embed graphics files
  \DeclareGraphicsExtensions{.pdf,.png,.jpg,.jpeg} % for pdflatex we expect .pdf, .png, or .jpg files
\else%                                 % else we use pure latex
  \usepackage{graphicx}                % allow us to embed graphics files
  \DeclareGraphicsExtensions{.eps}     % for pure latex we expect eps files
\fi%

\graphicspath{{figures/}{pictures/}{images/}{./}} % where to search for the images

\usepackage{microtype}                 % use micro-typography (slightly more compact, better to read)
\PassOptionsToPackage{warn}{textcomp}  % to address font issues with \textrightarrow
\usepackage{textcomp}                  % use better special symbols
\usepackage{mathptmx}                  % use matching math font
\usepackage{cite}                      % needed to automatically sort the references
\usepackage{tabu}                      % only used for the table example
\usepackage{booktabs}                  % only used for the table example
\usepackage{graphicx}
\usepackage{comment}
\usepackage{amsmath,amssymb} % define this before the line numbering.
\usepackage{color}
\usepackage{times}
\usepackage{epsfig}
\usepackage{amsmath}
\usepackage{amssymb}
\usepackage{dirtytalk}
\usepackage{dblfloatfix} 
\usepackage{siunitx} 
\usepackage{algorithm}
\usepackage[noend]{algpseudocode}
\usepackage{algorithmicx}
\usepackage{enumitem}

\usepackage[dvipsnames]{xcolor}
\usepackage{lipsum}

\usepackage{pifont}% http://ctan.org/pkg/pifont
\usepackage[breaklinks=true,bookmarks=false]{hyperref}

\newcommand{\tablestyle}[2]{\setlength{\tabcolsep}{#1}\renewcommand{\arraystretch}{#2}\centering\footnotesize}
\newlength\savewidth\newcommand\shline{\noalign{\global\savewidth\arrayrulewidth
  \global\arrayrulewidth 1pt}\hline\noalign{\global\arrayrulewidth\savewidth}}

\newcommand{\changed}[1]{{\color{black} #1}}

\usepackage{pdfpages}

\onlineid{1093}

\vgtccategory{Research}
\vgtcpapertype{Technology}

\title{Event-Based Near-Eye Gaze Tracking Beyond 10,000 Hz}

\author{Anastasios N. Angelopoulos*, Julien N.P. Martel*, Amit P. Kohli, J\"{o}rg Conradt, Gordon Wetzstein}
\authorfooter{
\item
Anastasios N. Angelopoulos and Amit P. Kohli are with the University of California Berkeley. E-mail: \{angelopoulos,apkohli\}@berkeley.edu,
\item
Julien N.P. Martel and Gordon Wetzstein are with Stanford University, E-mail: \{jnmartel,gordonwz\}@stanford.edu,
\item
J\"{o}rg Conradt is with the KTH Royal Institute of Technology. E-mail: jconradt@kth.se.
\item * denotes equal contribution
\item
\href{http://angelopoulos.ai/blog/posts/ebv-eye}{
Project website: \textcolor{blue}{http://angelopoulos.ai/blog/posts/ebv-eye}}
}

\shortauthortitle{Angelopoulos, Martel \MakeLowercase{\textit{et al.}}: Event-Based Near-Eye Gaze Tracking beyond 10,000 Hz}
\abstract{\changed{The cameras in modern gaze-tracking systems suffer from fundamental bandwidth and power limitations, constraining data acquisition speed to 300~Hz realistically. This obstructs the use of mobile eye trackers to perform, e.g., low latency predictive rendering, or to study quick and subtle eye motions like microsaccades using head-mounted devices in the wild.} Here, we propose a hybrid frame-event-based near-eye gaze tracking system offering update rates beyond 10,000~Hz with an accuracy that matches that of high-end desktop-mounted commercial trackers when evaluated in the same conditions. Our system, \changed{previewed in Figure~\ref{fig:teaser},} builds on emerging event cameras that simultaneously acquire regularly sampled frames and adaptively sampled events. We develop an online 2D pupil fitting method that updates a parametric model every one or few events. Moreover, we propose a polynomial regressor for estimating the \changed{point of gaze} from the parametric pupil model in real time. Using the first event-based gaze dataset, we demonstrate that our system achieves accuracies of 0.45$^\circ$--1.75$^\circ$ for fields of view from 45$^\circ$ to 98$^\circ$. With this technology, we hope to enable a new generation of ultra-low-latency gaze-contingent rendering and display techniques for virtual and augmented reality.%
} % end of abstract


\CCScatlist{ % not used in journal version
 \CCScat{K.6.1}{Management of Computing and Information Systems}%
{Project and People Management}{Life Cycle};
 \CCScat{K.7.m}{The Computing Profession}{Miscellaneous}{Ethics}
}

\teaser{
  \centering
  \includegraphics[width=\linewidth]{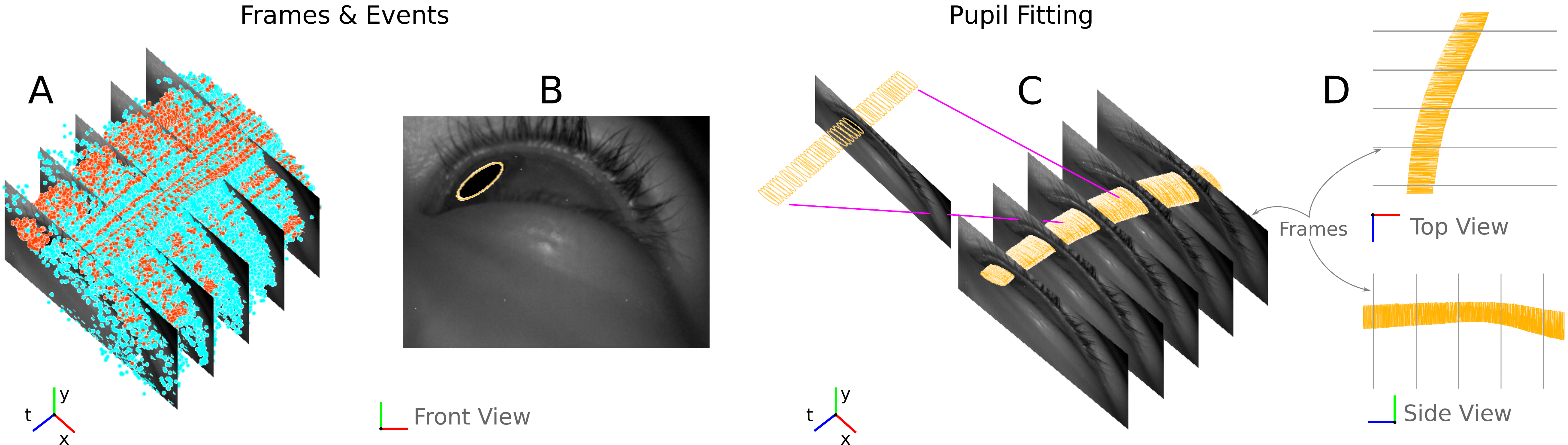}
  \caption{\textbf{Input and output of our system.}
	\changed{The inputs, shown in plot A}, are frames recorded at a fixed sampling rate and events, asynchronously sampling the eye motion at high speed. 
	Frames and events are captured by the same sensor, and event polarity is color coded in blue (+) or red (-). We output a \changed{gaze point}, computed from our estimate of the pupil, shown in yellow as seen from several perspectives \changed{in plots B, C, and D} (x and y are the columns and rows of the sensor, and t is time). \changed{Events continuously trigger between frames, allowing pupil estimation much faster than the frame rate. } Every pupil estimate yields a yellow circle. These estimates are so frequent that they form an almost continuous tubular structure outlining the pupil's movement through time \changed{in plots C and D}.
	}
	\label{fig:teaser}
}

\renewcommand{\manuscriptnotetxt}{}

\vgtcinsertpkg

\begin{document}


\firstsection{Introduction}

\maketitle
%%%%%%%%% BODY TEXT

\label{sec:intro}
\textit{Gaze tracking} is the process of estimating where a person is looking, usually with a camera. 
\changed{It enables dozens of applications in augmented and virtual reality (AR/VR), such as foveated or physiologically accurate rendering~\cite{guenter2012foveated,konrad2019gaze}, interactive programs that respond to eye movement by allowing the user to select a target with their eyes~\cite{kyto2018pinpointing}, and so forth.
Such applications, and others like laser eye surgery~\cite{neuhann2010lasik}, benefit from fast and accurate tracking of an eye that essentially fills the sensor field of view ('near-eye' tracking).
Ideally, to meet the battery and processing constraints of mobile headsets, eye trackers should also conserve power and compute.
}

%It enables dozens of applications in augmented and virtual reality (AR/VR) (e.g.,\cite{albert2017latency,patney2016towards,guenter2012foveated,padmanaban2017optimizing,konrad2019gaze}) but also in \changed{laser} eye surgery\cite{neuhann2010lasik}, target selection~\cite{kyto2018pinpointing}, user interaction~\cite{smith2013gaze,flickner2003method}, attention monitoring~\cite{smith2003determining,ishikawa2004driving,vicente2015driver}, psychological studies~\cite{mele2012gaze}, medical pathology~\cite{kundel2007mammogram}, and sports analysis~\cite{shank1987eye}. Fast and accurate near-eye tracking is desirable for most of these fields and crucial for AR/VR.

However, cameras force a tradeoff between resolution, framerate, and power, since \changed{every} pixel costs energy and bandwidth to acquire, communicate\changed{, and process}. Thus, the accuracy and latency of an eye tracking system are often in tension. High-end eye tracking systems resolve this using high-speed cameras, bespoke protocols, and customized readout interfaces to maximize bandwidth. Consequently, they are large and power hungry. These complexities are unavoidable because of the sheer volume of data generated by high-speed, high-resolution cameras. But in near-eye gaze tracking, most of this data is redundant. Only the pupil moves, while most of the image does not change. 

Dynamic vision sensors (DVS) overcome these limitations by adaptively sampling when the eye moves. DVS pixels separately sample when the instantaneous change in their incident irradiance exceeds a threshold. The result is a stream of pixel-by-pixel timestamped packets signaling changes, called \textit{events}. Events contain the pixel location, the time of sampling, as well as the sign of the change. In near-eye tracking, motion is sparse in time \textit{and} space. Events thus use bandwidth more efficiently than frames, because only relevant information is sampled and processed. This improves speed and power consumption.
%
%===========================================
\begin{figure*}[t!]
    \centering
    \includegraphics[width=\textwidth]{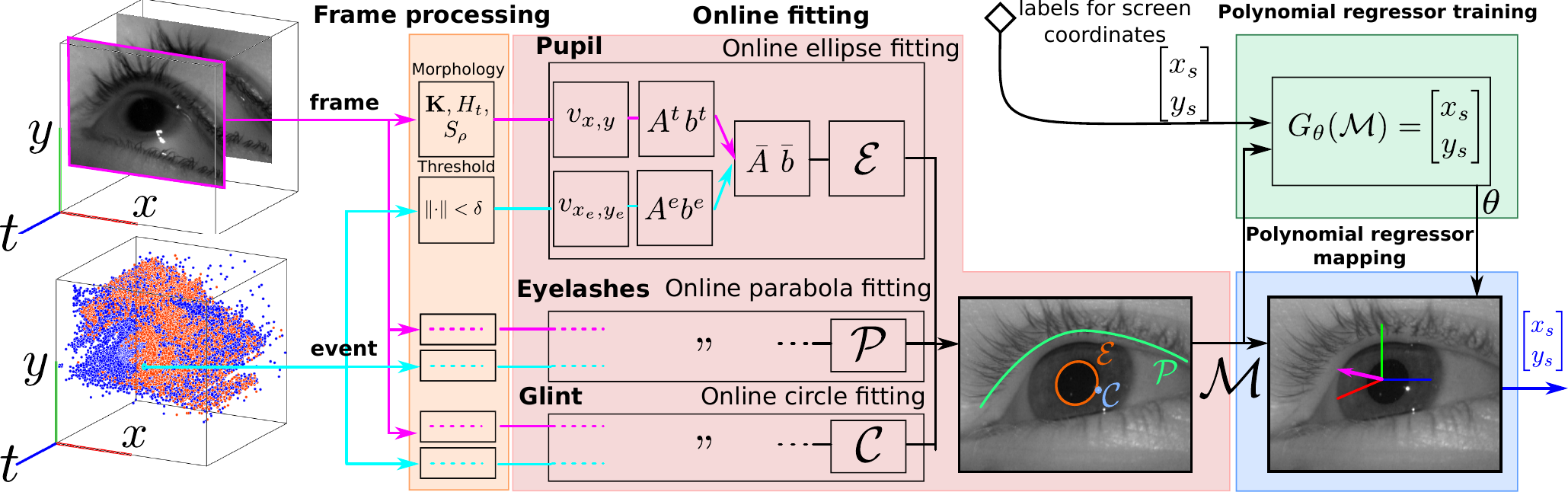}
    \caption{
    \textbf{A flow diagram of our system.} Both frames and events are inputs to our system. 
    A first stage preprocesses them separately before both streams are combined to update the eye model $\mathcal{M}=\{\mathcal{E},\mathcal{P},\mathcal{C}\}.$ \changed{Here, $\mathcal{E}$ and $\mathcal{P}$ are tuples of the parameters of quadratic functions defining the limbus and \changed{eyelid}, and $\mathcal{C}$ is a tuple of the two center coordinates (see Section~\ref{sec:methods} for further details).} The model fitting is performed online, in real-time at the arrival rate of frames (low, $~25~\si{\hertz}$) and events (high, up to $100~\si{\kilo\hertz}$). To output a gaze vector, $\mathcal{M}$ is fed to a $5$th-order polynomial regressor that can be evaluated at event rate. This regressor is trained via a calibration procedure similar to EyeLink's.
    }
    \label{fig:overall_system}
\end{figure*}
%===========================================
%

Here, we report the first real-time event-based eye tracking system, 
\changed{described in Figure~\ref{fig:overall_system}}. \changed{The sensor is placed close to the user's eye such that the eye nearly covers its field of view. For this reason, we refer to this system as being ``near-eye'', although our data is collected in a desktop setting. We discuss a miniature prototype in Section~\ref{sec:discussion}.} At the core of our system is the update of a parametric representation of the pupil at the event rate. This is fed to a polynomial regressor, also evaluated on an event-by-event basis, that maps this internal parametric representation to a \textit{gaze vector} (the 3D direction of gaze).
\changed{Along with events, we also use frames captured at low rates (15-20Hz) by the same sensor.}
%Along with events, we also use frames from an Active Pixel Sensor pixel sharing the same photodiode as the DVS. 
\changed{The frames anchor our pupil tracking system with traditional pupil-detection algorithms, while the events allow it to update the pupil's location at high-speed.}
In our system, gaze vectors can be queried at an estimated rate equivalent to $10,000~\si{\hertz}$ or more. This is an improvement of $>$10$\times$ over high-end, desktop-mounted devices \changed{(see Table~\ref{commercialtable})}using a small, low power sensor\footnote{A Pregius line 40FPS SONY sensor consumes 300mW--2W \cite{sonypregiusdatasheet} vs. 50mW--0.9W for a DAVIS-346 \cite{inivationdatasheet}.}. \changed{The following list summarizes our contributions.}

\begin{itemize}
    \item We introduce the first hybrid event-based eye tracking system and demonstrate a binocular prototype.
    \item We develop a model-based eye tracking algorithm, functioning at the event rate and implement it in a real-time system.
    \item We capture a binocular dataset of events and frames from 24 subjects performing saccadic motions and smooth pursuits.
\end{itemize}
\changed{
\subsection{Is 10kHz gaze tracking useful?}
    \label{subsec:10khz}
    The extreme speed of the eye tracking technology we introduce raises a natural question: why should we care about an extremely fast eye-tracking system?
    The answer has several layers.
    Firstly, the human eye moves quickly during certain motions, at speeds sometimes exceeding 300$^\circ / s$~\cite{verghese2017}.
    This fact justifies the use of very fast sampling, in the kHz range, in order to capture small, fast eye movements as they initiate and track their trajectories.
    Furthermore, the eye's acceleration regularly reaches astronomical values---e.g. 24,000$^\circ /s^2$---meaning the eye muscles are capable of making forceful and high frequency movements~\cite{abrams1989speed}.
    A simple Nyquist argument then motivates sampling in the 10~kHz range to fully capture the eye's motion.
    In particular, a sensor with a high peak sampling rate could detect when the eye is moving more quickly, characterize that motion with higher fidelity, and perhaps even enable predictive approaches to eye tracking.
    For applications like e-sports, where latency has a pernicious negative impact~\cite{kim2020post,kim2019esports}, fast event-based predictive rendering could be a {\em game-changer}.

    The adaptive sampling inherent to event-based cameras also helps speak to the utility of speed.
    Our system does not always run at 10~kHz, although it is capable of sustaining that rate.
    It only does so when the underlying motion of the eye generates many events.
    %The need for 10~kHz sampling is then somehow tautological---we did not decide to sample that fast, but rather, this rate naturally emerged from the motion of the eye.
    Rather than deciding a-priori to sample at 10~kHz, this rate emerged from the motion of the eye, a natural justification that this speed may be useful.
    This sampling approach allows the sensor to be much more power efficient than a bespoke camera system like the EyeLink 1000, which may not be suitable for an AR/VR headset due to power constraints.
}

\section{Related Work}
\label{sec:related}
\begin{table}
\centering
\tablestyle{2.5 pt}{1.05}
\begin{tabular}{ @{}l|c|c } 
 system & update rate (Hz) & accuracy ($^\circ$) \\ \shline
 Pupil Labs~\cite{pupillabsprice} & 200 & $\sim 1$  \\
 Tobii~\cite{tobiiprice} & 120 & 0.5--1.1  \\
 EyeLink~\cite{eyelinkinvoice,ehinger2019new} & 1,000 &  $\sim 0.5$  \\
 Ours  & $> 10,000$ & 0.45--1.75  \\
\end{tabular}
\caption{\textbf{Overview of existing eye tracking systems.} Our results can be directly compared with EyeLink because we use the same protocol \cite{eyelinkcommunication}. Our system's accuracy is $0.45^\circ$ within the same field of view as EyeLink, and speed is $10\times$ faster.}
\label{commercialtable}
\vspace{-0.5cm}
\end{table}
%

% \textbf{The concept of event based cameras}
\textbf{Event cameras} date back to the neuromorphic Silicon Retina introduced by the seminal works of Mahowald and Mead~\cite{mead1988silicon,mahowald1994silicon}, \changed{and have since advanced to a mature technology~\cite{delbruck1993silicon,lichtsteiner2008128,delbruck2010activity,posch2014retinomorphic}.}
In the intervening time, the utility of these high-speed sensors has been successfully exploited in a variety of application scenarios, including object and action tracking~\cite{mitrokhin2018event}, combined frame and event based object tracking~\cite{liu2016combined}, visual odometry~\cite{rebecq2017real}, 6-DOF pose tracking~\cite{mueggler2017poseestimation}, 3D reconstruction \cite{martel2018active},  SLAM~\cite{davison2007monoslam,weikersdorfer2014event}, and hand tracking~\cite{lee2012touchless}.
\changed{See the review by Gallego et al. for a dedicated overview of the diverse work in the event-based vision community~\cite{gallego2019event}.} 
%\textbf{Current gaze-tracking literature} generally focuses on one of two problems: gaze-tracking \say{in-the-wild,} on full-face images of users without infrared illumination, such as from a laptop webcam \cite{hansen2005eye}; or near-eye infrared-illuminated gaze-tracking for use in controlled environments like an AR/VR headset. We focus on the latter problem. We have summarized common commercial eye-tracking systems in Table \ref{commercialtable} \cite{pupillabsprice,ehinger2019new,tobiiprice,gibaldi2017evaluation,eyelinkinvoice}. No portable eye tracker achieves a framerate over 200~Hz.
%
\textbf{Hybrid frame-event approaches}
%In addition to the events, we build on prior work on event-frame fusion~\cite{delbruckfusion,gehrigfusion}, utilizing pupil estimates obtained at a low rate by an Active Pixel Sensor circuit placed on the same optical axis as the DVS pixel.
The fusion of events and frames has been explored in different works such as \cite{delbruckfusion,gehrigfusion}. In those approaches, direct, absolute photometric correlates acquired at low rate (the frames) initialize or enhance the estimation of a quantity tracked at high update rates (by the events). Similarly to visual--inertial odometry in which a high-rate, potentially drifting, sensor such as an inertial measurement unit (IMU) is corrected by a lower rate sensor that provides absolute anchor points (such as visual features), these fusion approaches are potentially more robust \cite{leutenegger2015keyframe}.

\textbf{Early approaches to eye tracking} include electro-oculography, search coils, and other invasive methods~\cite{young1975survey}.
Camera-based eye tracking has evolved from tracking Purkinje reflections~\cite{cornsweet1973accurate,crane1985generation} to model-based approaches that extract parameterizations of the eye from frames~\cite{tian2000dual,li2010eye,wang2017real}. The model extraction, or \textit{pupil fitting}, can be decoupled from regressing the \changed{point of gaze}. Algorithms for pupil fitting and gaze estimation are reviewed by Morimoto et al.~\cite{morimoto2005eye} and Duchowski et al.~\cite{duchowski2007eye}. However, unlike in event-based vision, these methods are rate-limited by camera frames.
A number of eye tracking strategies do not involve cameras, including photodiode base limbus trackers~\cite{topal2008head}, display embedded trackers for near-eye displays~\cite{vogel2009bi}, and LED based trackers~\cite{akcsit2020gaze,li2020optical}.
We do not focus our attention on these strategies, although they are promising.
%also, unlike appearance-based trackers, they have little shift invariance and are very sensitive to data quality.

%Cornsweet et al., using a custom optical setup which measured the Purkinje reflections of the eye, achieved an accuracy of 1' on an artificial eye, corresponding to $0.5\si{\degree}$ on real data, indicating that $0.5\si{\degree}$ may be the physiological limit of testing precision~\cite{cornsweet1973accurate,crane1985generation}. More recently, model-based approaches seek to extract explicit parameterizations of the eye from frames. The model extraction, sometimes called \say{pupil fitting} or \say{eye tracking}, can be decoupled from the regression. Multi-stage parametric model~\cite{tian2000dual}, deformable template matching~\cite{li2010eye}, and 3D models~\cite{wang2017real} are common approaches. Algorithms to extract eyes from frames and map them to \changed{points of gaze} are reviewed extensively by Morimoto et al.~\cite{morimoto2005eye} and Duchowski et al.~\cite{duchowski2007eye}. However, unlike in event-based vision, these methods are rate-limited by camera frames; also, unlike appearance-based trackers, they have little shift-invariance and are very sensitive to data quality.

\textbf{Appearance-based trackers} directly estimate \changed{points of gaze} from camera frames, often using neural networks~\cite{baluja1994non,krafka2016eye,ranjan2018light,zhu2017monocular}. A number of datasets have emerged over the past five years, many of which leverage synthetic data, including: GazeCapture~\cite{krafka2016eye}, UT Multi-view \cite{sugano2014learning}, SynthesEyes~\cite{wood2015rendering}, UnityEyes~\cite{wood2016learning}, MPIIGaze~\cite{zhang2019mpiigaze}, and, most recently, NVGaze~\cite{kim2019nvgaze}. These datasets comprise millions of synthesized and real eye images. However, there is no dataset for event-based eye tracking. 

%\textbf{Appearance-based trackers}, which treat images as points in a higher-dimensional space and map them to \changed{points of gaze}, have recently resurged in the computer vision community using deep learning. Artificial neural networks were first used in eye tracking by Baluja et al.~\cite{baluja1994non}, and have since been improved with CNN architectures~\cite{krafka2016eye,ranjan2018light,zhu2017monocular}. A number of datasets have emerged over the past five years, many of which leverage synthetic data, including: GazeCapture~\cite{krafka2016eye}, UT Multi-view \cite{sugano2014learning}, SynthesEyes~\cite{wood2015rendering}, UnityEyes~\cite{wood2016learning}, MPIIGaze~\cite{zhang2019mpiigaze}, and, most recently, NVGaze~\cite{kim2019nvgaze}. These datasets comprise millions of synthesized and real eye images. However, there is no dataset for event-based eye tracking. 

More recent gaze-tracking literature generally focuses on one of two problems: gaze-tracking \textit{in the wild}, i.e. on full-face images of users without infrared (IR) illumination, such as from a laptop webcam~\cite{hansen2005eye}; or near-eye IR-illuminated gaze-tracking for use in controlled environments like an AR/VR headset. \changed{The review by Koulieris et al. details these modern topics~\cite{koulieris2019near}.} Note that we do not focus on eye-tracking in the wild, although a deep learning model like NVGaze~\cite{kim2019nvgaze} used in the frame-based portion of our system might enable us to do so. Hence, our work falls in the second category, and we focus on eye tracking in controlled, near-eye settings. We evaluate against commercial eye-tracking systems in Table~\ref{commercialtable} using similar conditions and protocols. Our eye tracker, like all other infrared near-eye trackers, is not meant to be used where there will be large variations in head pose, reflections, or background. It is geared towards AR/VR or biomedical settings where high framerates are desirable (hence why the EyeLink 1000 samples at 1-2~$\si{\kilo\Hz}$). Consequently, the accuracy and update rate of our near-eye tracker with respect to controlled data is the proper evaluation metric. No portable eye tracker achieves a framerate over roughly 200~Hz; but using event-cameras, we achieve $>$10,000~Hz with comparable accuracy to the gold standard, desktop mounted EyeLink device. Compared to the EyeLink, our system is similarly accurate and an order of magnitude faster. Compared to mobile systems, our system is similarly accurate and two orders of magnitude faster.

%We build on the above work by creating an event-based tracker with high speed that takes advantage of recent advances in deep learning and providing a small but first-of-its kind hybrid event/frame dataset for the community to utilize. We use a DAVIS346 chip, harnessed by a custom-made system relying on ARM-core microprocessors. Example output of this system is depicted in Figure~\ref{fig:teaser}.
%

% discuss this last because it's the closest related work
The work closest to ours is the course project by Gisler~\cite{gisler2007eye}, who suggests the idea of using a DVS for eye-tracking. However, the accuracy of their system is limited to \textit{``a third of the size of a $1024\times 768$ screen''} and they did not demonstrate real-time performance. To the best of our knowledge, our work is the first to implement such a system, to develop a practical algorithmic framework that includes an online pupil fitting procedure, and to capture an event-based eye tracking dataset, which will be released to the public.

%The semester project of Gisler in 2007 $2007$~\cite{gisler2007eyetracking}, suggests the use of an event based sensor for eye-tracking, the resolution of their system is limited to a \say{third of the size of a $1024\times 768$ screen}, and does not demonstrate real-time event-based updates. To the best of our knowledge, our work is the first to implement such a system, to develop a practical algorithmic framework that includes an online pupil fitting procedure, and to capture a dataset, which will be released to the public.

\section{System and Methods}
\label{sec:methods}
%
%===========================================
\begin{figure}[t!]
    \centering
    \includegraphics[width=0.95\columnwidth]{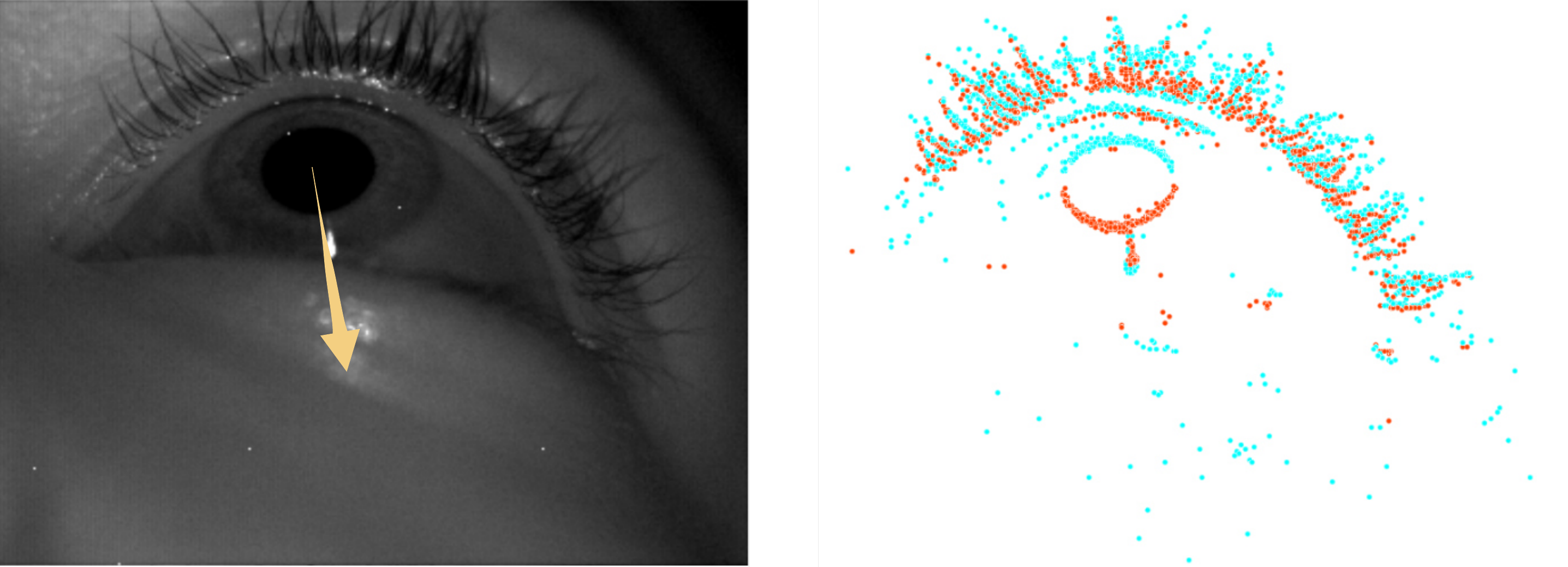}
    \caption{\textbf{Events are generated by the pupil as it moves.} The events on the right are generated by the eye milliseconds after the frame shown on the left, during a downward saccadic eye movement. The bottom generates negative (red) events and the top generates positive (blue) ones, since the pupil induces a negative contrast change in its direction of motion. Each event in the right panel is received sequentially as a separate packet.}
    \label{fig:explanatory}
\end{figure}
%===========================================
%

\begin{figure}[t!]
    \centering
    \includegraphics[width=\columnwidth]{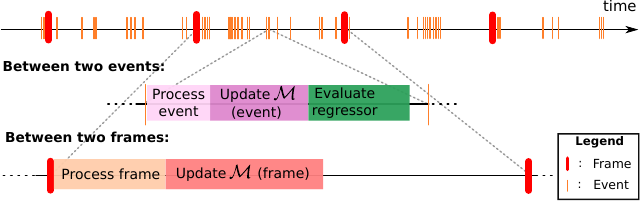}
    \caption{\textbf{The processing flow and time taken by the different stages in our system.} Our system allows events and frames to be processed concurrently and update the same underlying model. Events and frames are shown in time (top row). We also illustrate the operations happening sequentially and concurrently on events (middle) and frames (bottom).}
    \label{fig:chronology}
\end{figure}

A DVS pixel triggers independently from its neighbors when it sees a contrast change.
This is called an event. 
Events are like pixels in a difference image, except they are received separately at the time they were generated.
They carry no notion of a frame; events arrive as a sparse stream of data, \changed{as illustrated in Figure~\ref{fig:chronology}}. 
This stream consists only of the timestamps, signs ($\pm1$ for positive or negative contrast changes), and locations of the triggered pixels.
Sparsity comes with many benefits, like speed and processing efficiency.
However, previously developed frame-based eye tracking methods like binarization and dark/bright pupil tracking cannot be used on such data.
Our goal in this section is to develop a technique which can efficiently process this new data stream while still leveraging the large body of existing work in frame-based tracking.

To be precise, we built a hybrid event-frame-based gaze tracker combining the low latency of events with the proven robustness of frame-based methods. Our system consists of two distinct, concurrent processing stages. First, it fits a \textit{2D eye model} from the event and frame data. Second, it maps the estimated eye model parameters to a \textit{3D gaze vector} representing the direction the user is looking, or alternatively, a pixel on a screen at fixed distance (both representations have two free parameters). An overview of our algorithmic framework is illustrated in Figure~\ref{fig:overall_system}.

\subsection{2D Model Fitting}
\label{subsec:model_fitting}
%
%\textbf{Summary} 
We start by defining a parametric \textit{eye model}, with parameters $\mathcal{M}$. Whenever a frame or event is received from the sensor, these parameters are updated. Since frames are produced at a constant rate, they update the model independent of scene dynamics. But events only occur during eye motion, and update it at high frequency. Our method fuses these synchronous and asynchronous streams.

\textbf{Eye model} Motivated by Tian et al. \cite{tian2000dual}, our eye model consists of an \textit{ellipse representing the pupil} with parameters $\mathcal{E}=(a,h,b,g,f)\in\mathbb{R}^5$, a \textit{parabola representing the \changed{eyelid}} with parameters $\mathcal{P}=(u,v,w) \in\mathbb{R}^3$, and a \textit{circle representing the glint} (the reflection of the IR light source off of the user's eyeball) with parameters $\mathcal{C}=(r,c)\in\mathbb{R}^2$. The eye is thus fully parameterized by the 11 coefficients in  $\mathcal{M}=\{\mathcal{E},\mathcal{P},\mathcal{C}\}$. 

The parameters $\mathcal{E},\mathcal{P},\mathcal{C}$ are fit separately. Ellipses, parabolas, and circles are expressed canonically as quadrics, so they can be asynchronously estimated using the same method. In the following, we detail the updates for the fitting of $\mathcal{E}$ (see supplement for $\mathcal{P}$ and $\mathcal{C}$). The task ahead is to estimate $\mathcal{E}$ from a set of \textit{\say{candidate ellipse points}} $\mathcal{D}$ in the image plane that we believe lie on the edge of the pupil.

\textbf{Parameterizing the pupil with an ellipse} The locations $(x,y)$ of points on the ellipse representing the pupil in the image plane satisfy the quadric equation:
\begin{align}
    E_\mathcal{E}(x,y) &= 0 \\
    \text{with}\quad E_\mathcal{E}(x,y) &= a\,x^2 + h\,xy + b\,y^2 + g\,x + f\,y + d \nonumber
\end{align}
We set $d=-1$ for convenience as it is an arbitrary scaling factor corresponding to the offset of the plane intersecting the conic defined by $\mathcal{E}=(a,h,b,g,f)$. 

For each frame, we classify pixels near the edge of the pupil in an image as candidate points $\mathcal{D}_{\mathrm{img.}}$.  Events near the edge of the pupil are considered candidate points $\mathcal{D}_{\mathrm{evt.}}$. Thus, the model of the ellipse is ultimately updated by the set of points $\mathcal{D}=\mathcal{D}_{\mathrm{evt.}} \cup \mathcal{D}_{\mathrm{img.}}$.

\textbf{Receiving a frame} Under off-axis IR illumination, the pupil appears as a dark blob in the frame (see Fig. \ref{fig:setup}). By binarizing the greyscale frame $I$ using a constant threshold $\theta$, removing noise on the resulting image using morphological opening, and applying an edge detector, we identified the candidate points:
\begin{equation}
    \mathcal{D}_{\mathrm{img.}} = \left\{ (x,y) \;|\; \mathbf{K}\big(\mathrm{H}_{\theta}(I)\circ S_{\sigma}\big)(x,y)=1\right\},
    \label{eq:retrieve_dimg}
\end{equation}
where $\mathrm{H}_{\theta}$ is the \changed{unit step} function shifted by $\theta$ used for thresholding; $\circ$ denotes morphological opening; $S_\sigma$ is its structuring element, a discretized circle parameterized by its radius $\sigma$; and $\mathbf{K}$ is a binary edge detection function. We found that recovering candidate ellipse points with these simple operations worked sufficiently well. However, one could use any state-of-the-art frame-based pupil tracking algorithm outputting a set of candidate points to replace this stage of our system, such as PuReST \cite{santini2018purest}, ExCuSe \cite{fuhl2015excuse}, and Else \cite{fuhl2016else}. 

\textbf{Receiving an event} \changed{Events only contribute to the fitting of $\mathcal{E}$ when they are $\delta$-close to the border of the last estimated ellipse,
\begin{equation}
    \mathcal{D}_{\mathrm{evt.}} = \left\{ (x,y) \;|\;  \Vert P_{\mathcal{E}}\big( (x,y) \big)- (x,y) \Vert_2 < \delta \right\},
\end{equation}
where $P_{\mathcal{E}}\big((x,y)\big)$ is the projection of $(x,y)$ onto the ellipse --- this is step 8 of Algorithm~\ref{alg:fitting_of_evts_and_images}.

The projection operator $P_{\mathcal{E}}$ can be computed by solving a system of two equations (one linear and one quadratic) in the specific case of the ellipse parameterization.}

Our method can perform updates of $\mathcal{M}$ on an event-by-event basis, in which case $\mathcal{D}_{\mathrm{evt.}}$ is a singleton containing a single event. 
As we shall see in the experiments, the robustness of our method benefits from considering more than one event per update, in which case $\mathcal{D}_{\mathrm{evt.}}$ contains more than one event.
%
%
%===========================================
%===========================================
%

\textbf{Fitting the ellipse from images and events} We fit the ellipse model (and similarly the parabola and circle models) using least squares. The data points $\mathcal{D}_{img.}$ coming from the same frame can be thought of as having been generated synchronously, allowing us to fit the model to the data as a batch:
\begin{equation}
    \mathcal{E}^{*} = \underset{\mathcal{E} \in \mathbb{R}^5}{\arg \min} \sum_{(x,y)\in\mathcal{D}_{\mathrm{img.}}} E_{\mathcal{E}}(x,y)^2
\end{equation}
whose solution is simply $\mathcal{E}^* = A^{-1}\,b$ with
\begin{equation}
    A = \sum_{(x,y)\in\mathcal{D}_{\mathrm{img.}}} v_{x,y}\, v_{x,y}^\intercal,\quad 
     b = \sum_{(x,y)\in\mathcal{D}_{\mathrm{img.}}} v_{x,y} \label{eq:AandB_img} 
\end{equation}
and
\begin{equation}
    \quad v_{x,y} = (x^2, xy, y^2, x, y)^\intercal \label{eq:v_img}
\end{equation}
Generally, we start with an initial estimate of the ellipse's parameters and then wish to update it with new sensor information (such as a received frame or event). We can do so because in Equation~\eqref{eq:AandB_img}, $A$ and $b$ are sums and can thus be updated ``online." Practically, we store a matrix $\bar{A}$ and a vector $\bar{b}$, that are both updated upon the reception of new candidate points by ``summing them in." More formally, when a set of candidate points $\mathcal{D}^t$ arrives at time $t$, it is used to produce a matrix $A^t$ and a vector $b^t$ according to Equation~\eqref{eq:AandB_img}.  $A^t$ and $b^t$ can then be blended with a matrix $\bar{A}^{t}$ and a vector $\bar{b}^{t}$ representing and storing the ``current" state of the fit.
\begin{align}
    \bar{A}^{t+1} &= \gamma\,\bar{A}^t + (1-\gamma)\,A^{t},\;  \label{eq:update_e_from_img} \\ 
    \bar{b}^{t+1} &= \gamma\,\bar{b}^t + (1-\gamma)\,b^{t},\;\text{with}\, \gamma\in[0,1] \nonumber
\end{align}
%
%===========================================
\begin{algorithm}[t!]
\caption{ Online fitting of $\mathcal{E}$ from events and images}
\label{alg:fitting_of_evts_and_images}
\begin{algorithmic}[1]
\State $\bar{A} = \mathrm{Id}_{5\times5}$, $\bar{b} = \vec{0}_5$ \Comment{Init. $\bar{A}$ and $\bar{b}$}
\State $\bar{A}_{\mathrm{inv.}} = \bar{A}^{-1} = \mathrm{Id}_{5\times5}$ \Comment{If using SMW init. $\bar{A}_{\mathrm{inv.}}$}
\While{we are receiving data $d$}
    \State $\mathcal{D}_{\mathrm{img.}}, \mathcal{D}_{\mathrm{evt.}}  \gets \emptyset$
    \If{$d$ contains a frame $I^t$}
        \State $\mathcal{D}_{\mathrm{img.}} \gets \big\{ (x,y) \;|\; \mathbf{K}\big(\mathrm{H}_{\theta}(I^t)\circ S_{\rho}\big)(x,y)=1 \big\}$
    \EndIf
     \If{$d$ contains events with coordinates $(x_1,y_1), ..., (x_N,y_N)$}
        \State
        $\begin{aligned}
            \mathcal{D}_{\mathrm{evt.}} \gets \Big\{ (x_j, y_j) \; \lvert \; \Vert P_{\mathcal{E}}\big((x_j, y_j) \big)-(x_j, y_j) \Vert_2 < \delta,& \\  j \in \{1, ..., N\}& \Big\}
        \end{aligned}$
    \EndIf
    \ForAll{$(x,y)\in\mathcal{D}=\mathcal{D}_{\mathrm{evt.}} \cup \mathcal{D}_{\mathrm{img.}}$}
        \State $v_{x,y}^\intercal \gets (x^2, xy, y^2, x, y)$ \Comment{Eq. \eqref{eq:v_img}}
        \State $A \gets \sum_{(x,y)\in\mathcal{D}} v_{x,y}\, v_{x,y}^\intercal$ \Comment{Eq. \eqref{eq:AandB_img}}
        \State $b \gets \sum_{(x,y)\in\mathcal{D}} v_{x,y}$
    \EndFor
    
    \If{$|\mathcal{D}|>1$} \Comment{Batch update (frame/multiple evts.)}
        \State $\bar{A} \gets\; \gamma\,\bar{A} + (1-\gamma)\,A$ \Comment{Eq. \eqref{eq:update_e_from_img}}
        \State $\bar{b} \gets\; \gamma\,\bar{b} + (1-\gamma)\,b$
        \State $\bar{A}_\mathrm{inv.}\gets \bar{A}^{-1}$ \Comment{Full-rank inversion}
        \State $\mathcal{E}\gets\bar{A}_\mathrm{inv.}\,\bar{b}$
    \Else{} \Comment (Event-based update)
        \State $\bar{A}_\mathrm{inv.} \gets \frac{1}{\gamma'}\bar{A}_\mathrm{inv.} -
        \frac{1-\gamma'}{\gamma'}\frac{\bar{A}_\mathrm{inv.}\, A \,\bar{A}_\mathrm{inv.}}{\gamma'\, + (1-\gamma'\,)v_{x,y}^\intercal \,\bar{A}_\mathrm{inv.}\, v_{x,y} }$ \Comment{SMW}
        \State $\bar{b} \gets \gamma'\,\bar{b} + (1-\gamma'\,)b$
        \State $\mathcal{E}\gets\bar{A}_{\mathrm{inv.}}\,\bar{b}$
    \EndIf
\EndWhile
\end{algorithmic}
\end{algorithm}
%===========================================
%
This method has the advantage of storing a single small $5\times 5$ matrix and a $5$-dimensional vector, and blends information in time in a principled way. 
Our method is reminiscent of reweighted least-squares (RLS) with the loss of each candidate-point geometrically decayed by $\gamma$. However, it is not equivalent due to the thresholding operation. 
To prove this fact, in the setting of Equations 4--7, number $v_{x,y}^{(1)},...,v_{x,y}^{(t+1)}$.
RLS has a solution of the form $R_t^{-1}r_t$ where: $R_t=\Sigma_{i=1}^{t}a_iv_{x,y}^{(i)}v_{x,y}^{(i) \; T}$, $r_t = \Sigma_{i=1}^t a^{(i)}v^{(i)}$ and $a^{(i)}$ are constant discount factors. 
Given $\bar{A}^t$ and $\bar{b}^t$, we want $\bar{A}^{t+1}$ and $\bar{b}^{t+1}$ to have the RWLS form. 
But $A^{t+1}=\mathbf{1}(P_{\mathcal{E}_t}((x_{t+1},y_{t+1}))\leq t)(1-\gamma)(v_{x,y}^{(t+1)}v_{x,y}^{(t+1) \; T} - \bar{A}^t) + \bar{A}^t$, since $v_{x,y}^{t+1}$ is only included if it is close enough to the last ellipse.
The definition of $\bar{b}^{t+1}$ is analogous. 
Therefore Algorithm 1 is RLS if and only if $\mathbf{1}(P_{\mathcal{E}_t}((x_{t+1},y_{t+1}))\leq T$ is constant.
\changed{One can interpret $\delta$ and $\gamma$ as spatio-temporal regularization of our method; for example setting $\gamma=0$ will throw out all old candidate points (leading to degenerate ellipses because $A$ becomes rank-1) while setting $\gamma=1$ will stop all updates.}

In the case that $\mathcal{D}^t$ comes from a frame, $A^t$ and $b^t$ can be directly calculated from \eqref{eq:AandB_img} since $A^t$ is usually full rank. In contrast, events arrive one at a time and asynchronously. Since our goal is to take advantage of the low-latency and high-time resolution of the event generation process, we should update $\bar{A}$ and $\bar{b}$ from $\mathcal{D}_{\mathrm{evt.}}$, as often as every event. An event generates a single candidate point, but $v_{x_t,y_t}$ can nonetheless be computed using Equation~\eqref{eq:v_img}. The corresponding $A^t$ and $b^t$ for that event are computed using Equation~\eqref{eq:AandB_img}. Note that because $A^t$ is rank-1, it is not invertible. This is not surprising, since $\mathcal{E}$ has 5 parameters, therefore one needs 5 independent points to fit them. In case one aims at performing an update every $N$ events, $|\mathcal{D}_{\mathrm{evt.}}|=N$, and we can update $\mathcal{E}$ batch-wise, similarly to a frame.

Again, applying Equation~\eqref{eq:update_e_from_img}, we can update $\bar{A}$ as $\bar{A}^{t+1} = \gamma'\,A^t + (1-\gamma')\,\bar{A}^t,$ with $\gamma'\in[0,1]$. After the reception of the first 5 events in a non-degenerate configuration, $\bar{A}^t$ is rank-5 and can thus be inverted (given we keep blending in new information, it is generally invertible for the rest of time). Since $v_{x,y}$ and the blending of $A$ and $b$ are both easy to compute, these updates can be performed at the event rate in practice. But, updating $\mathcal{E}^*$ eventwise (typically up to 200 times per millisecond during a saccade) also entails computing $(\bar{A}^t)^{-1}$ eventwise, which might be computationally infeasible to perform in real time. However, because every event generates an $A^t$ that is rank-1, one can store $(\bar{A}^t)^{-1}$ and use the Sherman-Morrison-Woodbury (SMW) identity \cite{hager1989updating} to update it directly, online, after applying a small decay term to downweight old data in time. The fitting of the ellipse is summarized in Algorithm~\ref{alg:fitting_of_evts_and_images}. Again, the fitting of $\mathcal{P}$ and $\mathcal{C}$ is analogous. This formulation and implementation of least squares is well suited for the fusion of both the event and frame streams: it is a natural online method that is agnostic to the synchronicity of the data. 

\subsection{Mapping the eye-model to a \changed{point of gaze}}
\label{subsec:mapping}
The output of the gaze tracker \changed{is the 2D screen coordinate the user is looking at, which we call the point of gaze}. In the first stage of our system (Sec.~\ref{subsec:model_fitting}), we fit the parameters $\mathcal{M}=\{\mathcal{E},\mathcal{P},\mathcal{C}\}$ of an eye model given incoming events and frames. We now discuss how we associate a \changed{point of gaze} to those parameters.

The 2D screen coordinate position a user is looking at is denoted $(x_s,y_s)$. The problem is to find a mapping from $\mathcal{M}$ to $(x_s,y_s)$. We could explicitly model and fit the relative poses of the camera, user eye, and screen along with the projection of the camera and transformation between screen and world coordinates. \changed{However, we adopt another approach, common in the gaze mapping literature \cite{morimoto2005eye} that consists of phenomenologically regressing the output $(x_s,y_s)$ from the pupil center $(x_c,y_c)$ using two 5$^{\textrm{th}}$ order polynomial functions $G_{\theta^1}|_x$ and $G_{\theta^2}|_y$ (one for each coordinate)}:
%
%===========================================

\begin{figure}[t!]
%\begin{SCfigure}
    \centering
    \includegraphics[width=\columnwidth]{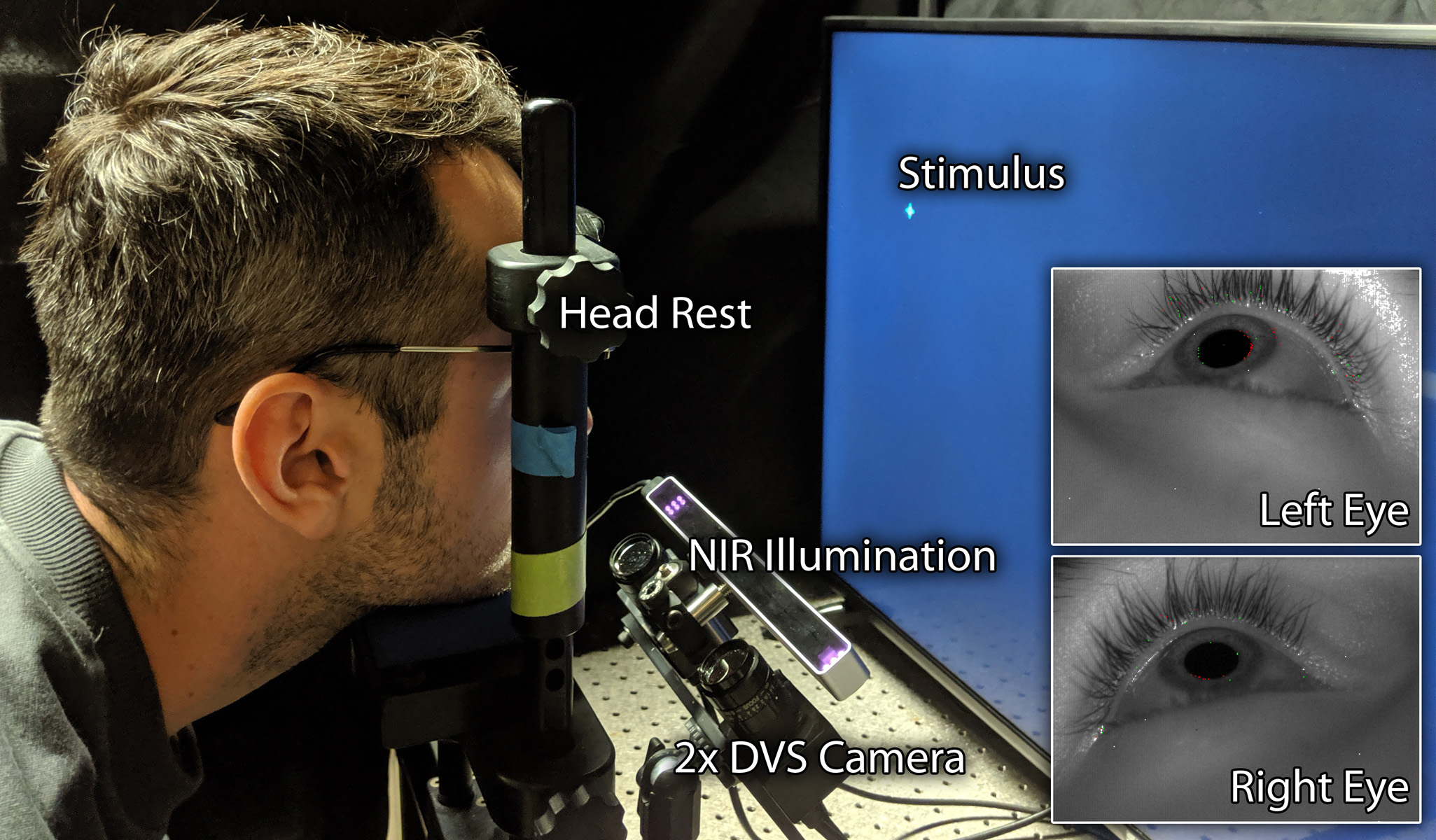}
    \caption{\textbf{The binocular eye tracking system setup used for evaluation.} We collected a dataset with binocular saccadic and smooth-pursuit data on 24 subjects looking at an 11x11 grid of fixation points over a $64 \times 96^\circ$ FoV. Altogether, we collected $\sim 30$ million events per subject per eye. Two DAVIS sensors and a near-infrared illumination source are mounted close to the user's head. The head is fixed by a head rest and the user observes a stimulus on the screen.}
    \label{fig:setup}
\end{figure}
%\end{SCfigure}
%===========================================
\changed{
\begin{equation}
    G_\theta(x_c,y_c) = \begin{pmatrix}
                       x_s \\
                       y_s
                     \end{pmatrix}
     = \begin{pmatrix}
       G_{\theta^1}|_x(x_c,y_c)\\
       G_{\theta^2}|_y(x_c,y_c) 
     \end{pmatrix} 
\end{equation}
}
%===========================================
\begin{figure*}[t!]
    \centering
    \includegraphics[width=\textwidth]{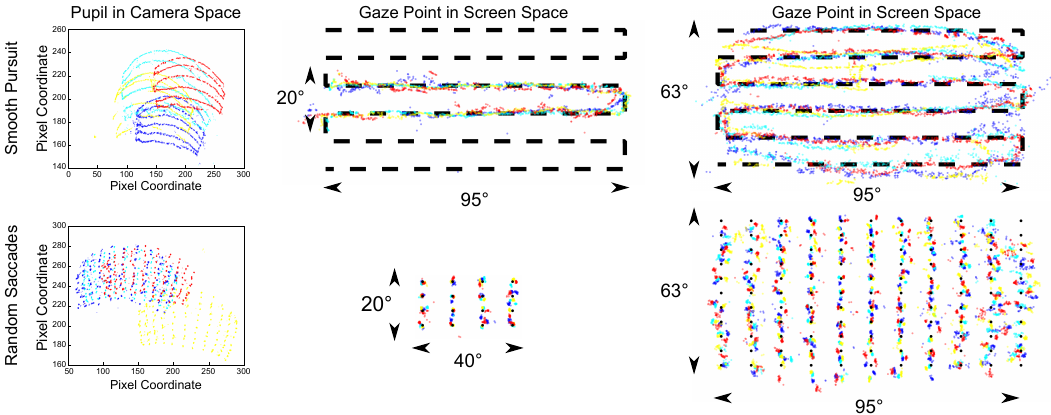}
    \caption{\textbf{Fitted pupil locations and gaze point estimates for smooth pursuit motion and random saccadic motion} are shown for four different users in different colors.
    \changed{The figure is organized into grids; the first row plots smooth pursuit data and the second row plots random saccadic data. The first column shows the extracted pupil center in camera image space, and the second/third columns shows the gaze point in screen space for a small/large field of view, with the ground truth locations indicated in black.} The average visual angle accuracy in a 47$\si{\degree}$ FoV for smooth pursuit data is 1$\si{\degree}$ (top center) and it is 3.9$\si{\degree}$ for the entire $113\si{\degree}$ FoV (top right).} 
    \label{fig:dataset}
\end{figure*}
%===========================================
%===========================================
\begin{figure*}[t!]
    \centering
    \includegraphics[width=\textwidth]{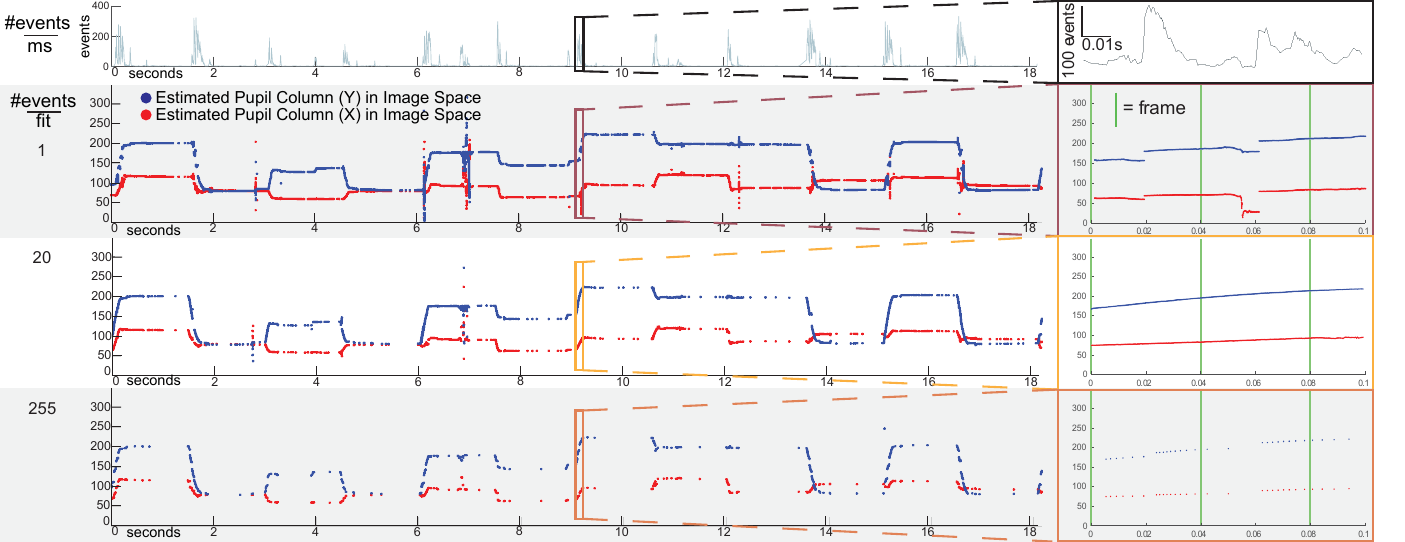}
    \caption{ \textbf{The X-Y coordinates of the fitted pupil over time.} Our system can update the pupil's location as fast as every event (row 2), but by considering more events, we can improve robustness and sparsity (row 3). Updating too slowly, however, harms robustness again (row 4). In the top row, a plot of the number of events per millisecond is shown for one user. Then, from top to bottom, different plots of the X (in red) and Y (blue) coordinates of the pupil position tracks are shown with an increasing the number of events used for each fit. One pupil position is obtained per fit. Hence, fewer fits are produced in the lower plot (as more events are used per fit) than in the upper plots. The number of fits also varies with the number of events produced per millisecond (top plot). We find $20$ events per fit to give a good trade-off between the speed of our system (the maximal number of fits per second we can obtain) and the robustness of the fit. The $10~\si{\kilo\hertz}$ peak update rate we report is calculated for $20$ events and based on the typical $200$ events we observe per millisecond during a saccade. The ``glitches" at 3 and 7 seconds are blinks, from which the system recovers at the next frame.}
    \label{fig:speed}
\end{figure*}
%===========================================
%
%===========================================
\begin{figure*}[t!]
    \centering
    \includegraphics[width=0.9\textwidth]{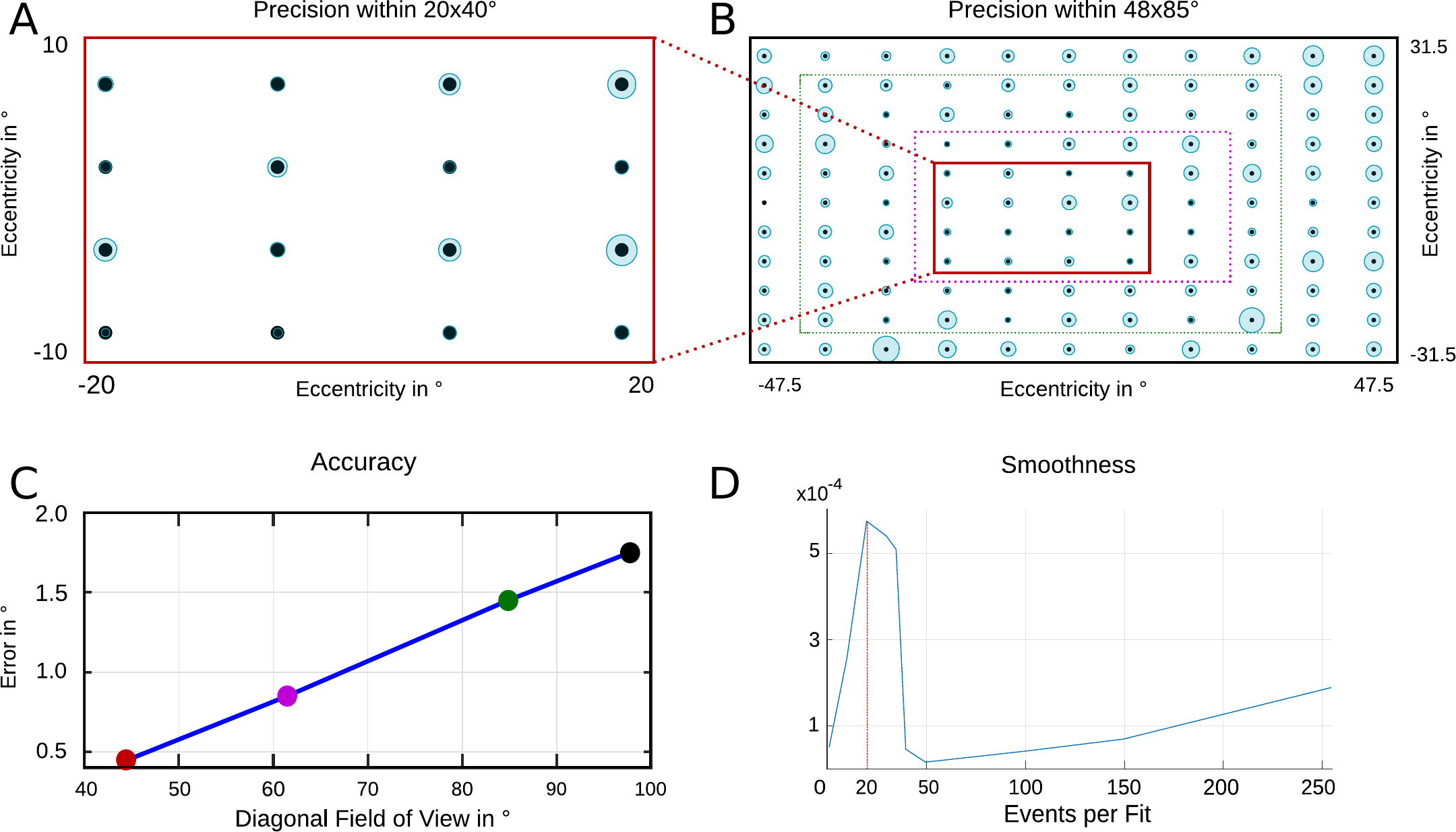}
    \caption{\textbf{Precision and accuracy of our system} depend on the target field of view. 
    The circle diameters \changed{in plots A and B} represent the precision, defined as the trimmed standard deviation of estimated gaze points centered at their ground truth label, averaged across all subjects.
    \changed{ Plot A is a restricted field of view from plot B.}
    The average precision is 1.6$^\circ$ in the smaller field of view and 3$^\circ$ in the larger. 
    Calibration is performed using only half the points within the evaluated field of view. 
    The black circles in the center represent the stimulus size, 1.3$^\circ$. 
    \changed{Plot C} shows the best accuracy of a single subject, which ranges from 0.45--1.75$^\circ$ for diagonal fields of view between 45--98$^\circ$. 
    \changed{Plot D illustrates} the \changed{empirical smoothness (see the Appendix for rigorous definition)}, against the number of events per fit for our method. 
    The plot shows a clear optimum at around $20$ events per fit, indicating that the robustness is highest for this setting}
    \label{fig:accuracysmoothness}
\end{figure*}
%===========================================
%
The simplicity of the model and polynomial regressor is a strength of our system. An advantage of using such a polynomial $G_{\theta}$ to map an eye model to screen coordinates is that it requires virtually no computation to evaluate: a few additions and multiplications. Hence, it is particularly well suited for event updates, even at high rates. Regressing the parameters $\theta$ of $G_{\theta}$ requires \changed{$\left((x_c,y_c), (x_s,y_s)\right)$ input-output training pairs} (the exact number depending on the exact degree of the polynomial) which are generally obtained during calibration by extracting the pupil center as in Section~\ref{subsec:model_fitting} and regressing against the known gaze point. \changed{Other simple regression strategies that we did not attempt, like Gaussian processes and random forests, may provide further benefits like uncertainty quantification.} Although a more complex model such as NVGaze \cite{kim2019nvgaze} may improve robustness and generalization, it requires orders of magnitude more computation and power and does not provide an accuracy advantage (see discussion).
%
%However, regressing the parameters $\theta$ of $G_{\theta}$ requires many $\left(\mathcal{M}, (x_s,y_s)\right)$ input-output training pairs (the exact number obviously depending on the exact degree of the polynomial). This is achieved through calibration procedure with a user looking at a-priori known position $(x_s,y_s)$ on a screen and using the fitted eye models $\mathcal{M}$ for each of those. Indeed, since the mapping $G$ depends on the relative position of the user to the screen, of the relative position of the camera to the eye, this needs to be done for each user individually, and might eventually have to be redone during the operation of the system in case the user moves.
%

\section{Dataset}
\label{sec:data}

\textbf{Data Acquisition} 
Our setup is shown in Figure~\ref{fig:setup}. It consists of two DAVIS346b (iniVation) sensors, imaging the right and left eye of a user placed at about $25~\si{\centi\meter}$ from each camera center. The user's head is fixed on an ophthalmic head rest and strapped during the experiment to prevent excessive slippage. The sensors are mounted with two $25~\si{\milli\meter}$ f/$1.4$ VIS-NIR C-mount lenses (EO-\#67-715) equipped with two UV/VIS cut-off filters (EO-\#89-834). The eyes are illuminated using the NIR illuminators of the Eye Tribe tracker. Both sensors are synchronized (their timestamps are aligned), and the $8$-bit $346\times 260$~px greyscale frames and events of both DAVIS346b sensors are recorded simultaneously. The exposure is set so as to maximize contrast in the frame, resulting in a frame rate of about $25$~FPS. The stimuli are displayed on a $40$~in diagonal, $1920\times 1080$~px monitor (Sceptre 1080p X415BV\_FSR), placed $40~\si{\centi\meter}$ away (standard reading distance) from the user \cite{dexl2010device,legge1999measuring}. It is horizontally centered and vertically aligned so that a user looking straight roughly gazes at a point placed at a third from the top.

\textbf{Dataset Characteristics}
Our dataset is, to the best of our knowledge, the first collected for gaze-tracking using event-based vision sensors. It was recorded on $24$ subjects and consists of two experiments corresponding to two different types of eye-motion: random saccades and smooth pursuit. The stimulus is a $40\times 40$~px green cross centered on a $20$~px diameter disk presented against a black background. In our setup the monitor spans a field of view (FoV) of $64 \times 96\si{\degree}$.

In the first experiment, users were asked to fixate on the stimulus randomly displayed at one of $121$ different locations (corresponding to an $11\times 11$ grid on the monitor) for $1.5~\si{\second}$ each. All locations are presented once, and the random sequence was the same for all users. The grid is visualized in the bottom row of Figure~\ref{fig:dataset}. In the second experiment, users were asked to fixate on the stimulus, which moved smoothly along a predictable square-wave trajectory starting at the top of the screen and moving towards the bottom while spanning the whole screen horizontally with a vertical period of $150$~px. This trajectory is the black dotted line in the top row of Figure~\ref{fig:dataset}.
\changed{
Although we only explicitly induced saccadic and smooth pursuit motions, our dataset contains rich eye motions including microsaccades and tremor. 
As one example, carefully parse the top row of Figure~\ref{fig:speed}, and notice that after every saccade (large spike in events) there is a corrective microsaccade (small spike in events).
Careful inspection of our public dataset will reveal detailed information about such subtle eye motions (e.g., the time between a saccade and the subsequent microsaccade).
}

\section{Results}
\label{sec:results}
%\textbf{Tracking the eye beyond 10,000 Hz: smoothness, sparsity, and update rate}
%
%===========================================

\textbf{Calibration} (i.e. defining $G_\theta$ and estimating $\theta$) is addressed in Section~\ref{sec:methods} and with more details in Supplement S2 and S4. Before using our eye tracker, a user looks at a set of ``calibration points'' whose coordinates in screen space are known. The pupil position in camera space is then extracted for each point, and a second-order polynomial is regressed mapping the pupil center to the screen coordinates: this is the \changed{gaze point}. Anytime we report accuracy or precision results for a particular FoV, we only calibrate on half the points in that FoV (e.g. odd indexes). Then, we report results on the full set of points. For a similar FoV to EyeLink's, the center $20^\circ\times40^\circ$ FoV, this means we are testing on 16 points and calibrating on 8. Eyelink uses a 12 point calibration procedure~\cite{eyelinkcommunication} which supposedly enhances accuracy. Our results can thus be readily compared to the commercial gold standard.

\textbf{Assessing Update Rates}
Our system can operate in real-time and update the pupil fit on an event-per-event basis; a conservative estimate of the peak update rate of our system is 10,000 Hz. We can achieve such a rate because saccadic motions induce hundreds of thousands of events due to the high-contrast edge between the pupil and the iris. These events can thus be used to \say{track} the pupil between two frames (see Figs.~\ref{fig:teaser},\ref{fig:speed}). In contrast, when the eye is still, very few events are produced.

We calculate the update rate as follows: we first estimate, in a conservative way, the amount of events per second for a saccade $\rho$. Figure~\ref{fig:speed} shows a saccade typically induces more than $\rho=200~\mathrm{evts}.\si{\milli\second}^{-1}$. Second, we calculate the optimal number of events per fit $N$ to produce a robust (smooth) pupil position estimate: this is $N^*=20$ according to Figure~\ref{fig:accuracysmoothness} in which we have performed an experiment varying the number of events per fit. This yields the conservative update rate of our system, $R=\frac{\rho}{N^*}$ which is $R=\frac{200\cdot10^{-3}}{20}=10~\si{\kilo\hertz}$. Our system can sustain an update rate of 10,000 Hz or above indefinitely, but this is not desirable because when the eye is still, no updating is required. Our event-driven update rate therefore only samples quickly when the motion of the eye requires it. In other words, there would be no speed advantage to using a frame-based system running at $=10~\si{\kilo\hertz}$.

The number of events used to perform a fit is the number of events accumulated in $\mathcal{D}$ before solving for $\mathcal{E}$ (in Algorithm~\ref{alg:fitting_of_evts_and_images}). Figure~\ref{fig:speed} illustrates the use of different values of $N$: we plot the fitted pupil center coordinates in image space for a random subject performing the random saccade experiment. As expected, when every event is used to update $\mathcal{E}$, the update rate is very high ($N=1$, $\rho=200~\mathrm{evts}.\si{\second}^{-1}$ thus $R=200~\si{\kilo\hertz}$) but the algorithm is not robust to series of noisy events which cause the fit to change drastically and reach an unrecoverable state until the next frame corrects it. Moreover, fitting for every event causes us to make wasteful updates to $\mathcal{E}$ even when the eye is not really moving and events are just noise. In the opposite case, where we consider $N$ to be hundreds, our data is desirably sparse and very robust (even to large perturbations such as blinks), but it does not smoothly follow a saccade between frames. This exposes an inherent tradeoff in our system between the smoothness, sparsity, and update rate.

In order to find an optimal value for this tradeoff, Figure~\ref{fig:accuracysmoothness} plots a measure of smoothness against the number of events per fit for a given subject. The quantitative measure of smoothness we use is the inverse norm of the concatenated 1-forward differences of the X and Y coordinates. It shows that a clear optimal is obtained for this measure of smoothness using $N=20$ events per fit. Indeed, this is also confirmed visually in the middle row of Figure~\ref{fig:speed}: this parameter has a good balance between sparsity, robustness, and speed.

\begin{figure}[t!]
    \centering
    \hspace{-0.25cm}\includegraphics[width=0.5\textwidth]{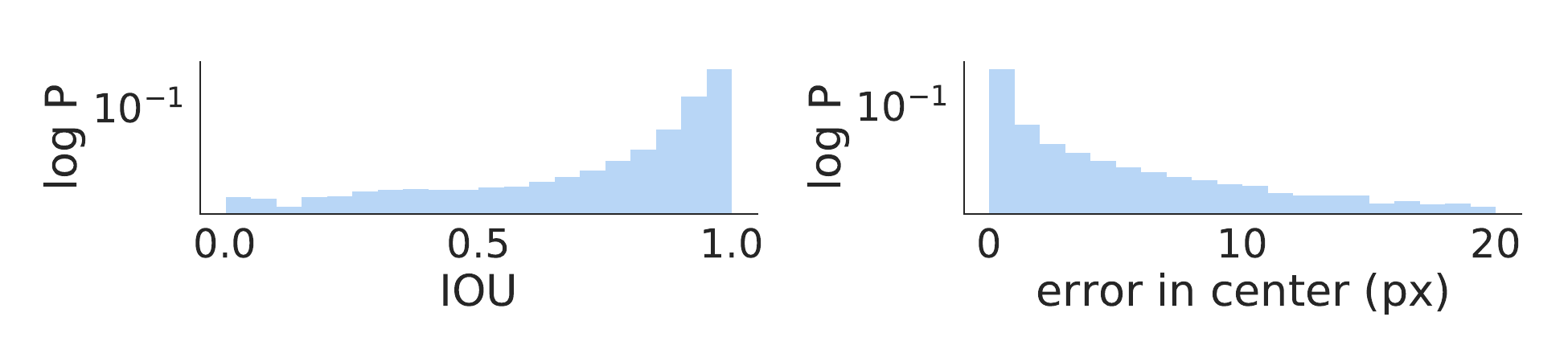}
    \vspace{-0.7cm}
    \caption{\textbf{The error in center estimation and intersection over union (IOU)} \changed{of our method over the full field of view and all subjects are plotted as normalized histograms. Notice the log scale on the y axis.}}
    \vspace{-0.5cm}
    \label{fig:pupil-metrics}
\end{figure}

\changed{
\textbf{Pupil Tracking Evaluations}
To assess the effectiveness of event-based pupil tracking, we plot the intersection over union (IOU) and error in center estimation as histograms in Figure~\ref{fig:pupil-metrics} over our entire dataset.
For two binary matrices $x$ and $y$, IOU can be expressed as 
\begin{equation*}
    IOU(x,y) = \frac{\sum x \lor y}{\sum x \land y },
\end{equation*}
where the logical operations are performed elementwise.
In our case, $x_{i,j}=1$ if the pixel $(i,j)$ lies within the event-based pupil estimate.
The matrix $y_{i,j}$ corresponds similarly to the ground truth pupil extracted from a frame.
IOU measures the quality of the pupil estimate $x$ an IOU of $1$ implies perfect overlap of $x$ and the ground truth $y$.
We compared the event-based pupil estimate obtained immediately before every frame in our dataset and compared it to the the pupil estimate from the frame.
The IOU almost never falls below 0.8 and the pupil center almost never deviates by more than 3 pixels.
These results inspire confidence that our method tracks the true underlying motion of the eye, since the events agree with an independently estimated pupil from the future frame. 
}
%===========================================
%===========================================

\textbf{\changed{Gaze Mapping} Accuracy}
Accuracy represents the closeness of the pupil estimates to ground truth. Specifically, we calculate it as the average absolute deviation of the pupil estimates from the labelled gaze points in the random saccade experiment. Our system achieves an accuracy comparable with commercial frame-based eye tracking systems, $<0.5^\circ$ \cite{ehinger2019new}, within a standard field of view, which degrades to $2^\circ$ on a larger field (see bottom row of Figure~\ref{fig:accuracysmoothness}). 
A purely frame-based eye-tracking system provides a lower-bound on our accuracy since events add information during fast motions of the eye that frames cannot sample. However, there is no way to evaluate the accuracy of event-wise updates of the pupil's position with a traditional camera and monitor. Hence, we performed our own experiments, in which we compared our frame-based algorithm's estimated gaze-locations with the ground truth point a user was looking at on a screen during saccadic motions. 

%===========================================
\begin{figure*}[t!]
    \centering
    \includegraphics[width=0.95\textwidth]{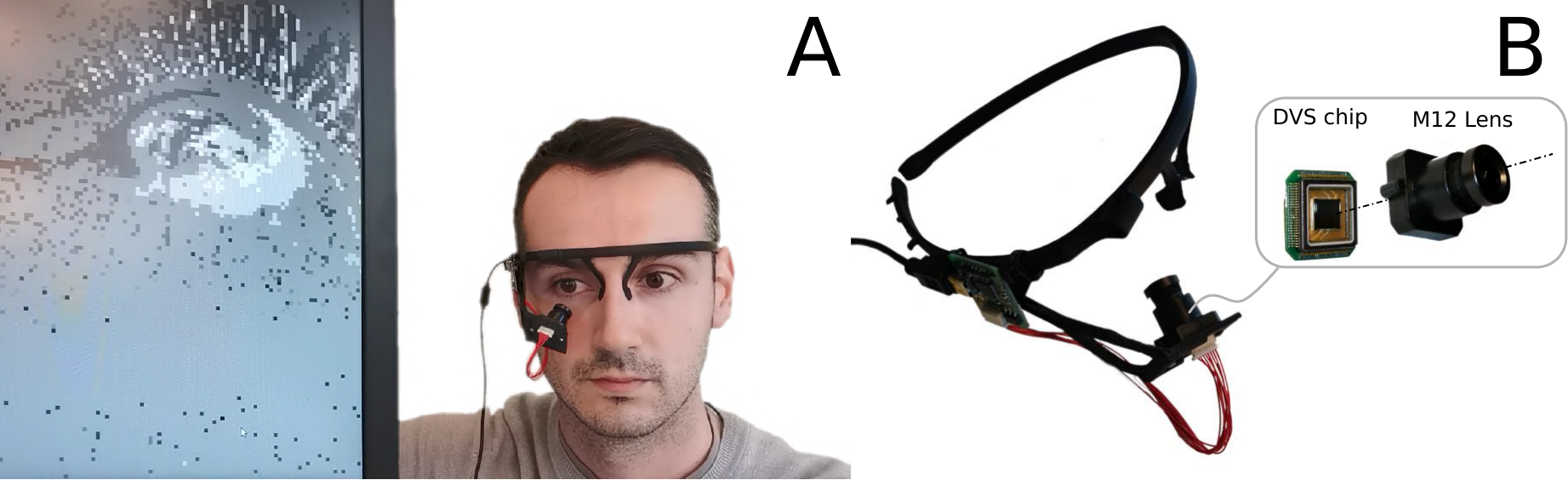}
    \caption{\textbf{Our miniature prototype} \changed{streams events in real time (A) and mounts on a pair of glasses (B), shown here with a M12 lens. Data is streamed out of the prototype using less than $12$Mbits/s of bandwidth.}}
    \label{fig:sensor}
\end{figure*}
%===========================================
Although we cannot directly evaluate the accuracy in-between frames, using the assumption that eye motion is continuous in camera-space at a very small timescale (this is why we optimized for smoothness of the trace in Figure~\ref{fig:speed}), we can indirectly assess accuracy at event rate. If our event-based updates did not match the motion of the pupil, then, when a frame was received, the X and Y pupil center traces in Figure~\ref{fig:speed} would have a \say{glitch} corresponding to the correction of the bad estimate (as in the case of the blink at roughly the 7 seconds mark). This does not occur, so our system must at least match the accuracy of our frame-based method. 
%One corollary of our result is that instead of using a frame-based tracker with extremely high frequency, such as the EyeLink 1000, it may be more efficient and equally accurate to use a hybrid event and frame based system with frames taken at a much slower speed, as we show in Figure~\ref{fig:speed}.
%\textcolor{red}{NOTE: As further evidence, provide an example of a saccade which starts before a frame, goes thru a frame, and ends after a frame. This might have to go into Figure~\ref{fig:speed}.}

\textbf{\changed{Gaze Mapping} Precision}
We calculate precision as the \changed{empirical} standard deviation, in visual angle, of the estimated pupil centers and plot it in \changed{Figure~\ref{fig:accuracysmoothness}A and B} across all subjects. \changed{The blue circles in the plot are thus a measure of the statistical spread, while the accuracy is reported as a line in the Figure~\ref{fig:accuracysmoothness}C.} We assess precision by fitting a polynomial on a subset of points and evaluating on a held-out set. Specifically we train on all the ``even" positions on the grid and evaluate on the ``odd" ones, do the opposite and average the numbers. We obtain $1.6\si{\degree}$ of visual angle precision on the smaller field of view. This precision decreases on the larger field to $3\si{\degree}$. Precision can also be visually assessed in Figure~\ref{fig:dataset}. In Figure~\ref{fig:accuracysmoothness}, only the last half second (out of a $1.5~\si{\second}$ stimulus presentation) is used for each saccade, as we assume the user's gaze might have changed in the first second (due to the large FoV, some subjects had to search for the stimulus). Blinks are removed using an automated blink detector which is part of our pipeline (see supplement). The top and bottom center plots are obtained by fitting a second-degree polynomial regressor on a smaller FoV, while the right plots are obtained by fitting a second-degree polynomial on a larger FoV.
\changed{Notice that at the edges of the top right and bottom left displays of Figure~\ref{fig:accuracysmoothness}, we can clearly observe that the accuracy and precision both worsen in the edges of the field of view, both because of occlusion and also because a small change in pupil center can have a large effect on gaze location when viewed at that angle.}  
%
% in which subjects were hand-picked to reflect low/medium precision ($1\si{\degree}$ on a smaller FoV and $2\si{\degree}$ of error in a larger FoV) and high errors ($3$ and $7\si{\degree}$ for the smooth pursuit data (top row).
%
%===========================================

We quantitatively assess the precision of our system in Figure~\ref{fig:accuracysmoothness}. A second-degree polynomial is fit on the best-subject for both a small $20\times 40\si{\degree}$ (top left) and larger $63\times 95\si{\degree}$ (top right) field of view. The total FoV $64\times 96 \si{\degree}$ spanned by the screen in our experiments is comparatively much larger (and harder to regress) than previously reported in the literature ($26\times 40\si{\degree}$ etc.) \cite{ehinger2019new}, explaining why we report results for two regressors fitted on two FoV. Precision in visual angle averaged for all the points in the small FoV is $1.6\si{\degree}$.

\section{Discussion}
\label{sec:discussion}
We presented a system and a method for near-eye gaze tracking using a vision sensor that can produce both conventional images and event data. Our system inherits the capabilities of frame-based sensors: we demonstrated state-of-the-art precision of $0.5\si{\degree}$ of visual angle error in a $121$ fixation point task, in addition to the advantages of event-based data. Specifically, we obtain a conservative peak rate of $200$ events per millisecond in a $2$~px radius around the pupil and showed our method can achieve a robust fit performed every $20$ events. Hence we claim a conservative peak update rate of $10~\si{\kilo\hertz}$. Our method can sustain the real-time processing of those $200~\mathrm{evts}.\si{\milli\second}^{-1}$.  A method that could reliably estimate the pupil position every single event, or would be able to generate/consider an even higher event rate, would theoretically yield even higher update rates. The update rate we demonstrate is about $10\times$ higher than the fastest commercial systems we surveyed, and such cannot be envisioned, even with a modest resolution conventional camera, due to the bandwidth required to output frames at those frequencies.

Limitations of our work, as well as most other existing eye trackers, include susceptibility to slippage and unavailability of ground truth. Specifically, $G_\theta$ is not robust to slippage of the cameras with respect to the face. This is an open challenge that, for now, requires periodic recalibration. Ground truth is unavailable for all eye-trackers; the standard method of evaluation is to assess both against one-another with the same methodology~\cite{eyelinkcommunication}. A final limitation is that we do not report the time between the true movement of the human eye and the time we register an update, otherwise known as latency. Latency is heavily dependent on the hardware details of the processor and the communication channel with the camera. Therefore, reporting latency is outside the scope of our work, since we did not attempt to build a custom embedded system. However, due to our extremely fast update rate, our software would surely not limit the speed of a commercial system, and optimizing the hardware would ensure fast operation. We estimate each event based update to require, in worst case, about 300 FLOPS. This means our algorithm lends itself to dedicated implementations that are likely to be very efficient on most modern low-power embedded-processors. 
Additionally, although it was not the focus of our work, there is the possibility of using a deep convolutional neural network to output a pupil center estimate in our gaze tracking pipeline (see Figure~\ref{fig:overall_system}, top). This may improve the robustness of our system and allow generalization to different subjects, like NVGaze \cite{kim2019nvgaze}. It is worth noting, however, that the best-case accuracy NVGaze reports when trained and evaluated on a single subject in a near-eye scenario is $0.5^{\circ}$, the same as ours. Additionally, using such models in a compute-constrained setting may bottleneck tracking speed or consume too much power.
\changed{
Finally, our choices of the tuning parameters $\gamma$, $\delta$, and \# of events per fit (see Figure~\ref{fig:accuracysmoothness}) depend on assumptions about the true Lipschitz parameter of the eye's motion.
Partially because the true Lipschitz parameter is unknown, changing, and perhaps unknowable, the user of our system will need to hand tune these parameters to strike a balance between the sensitivity of the method to jerky eye motions and its susceptibility to noise.
Because our dataset consists mostly of smooth motions like fixation, saccades, and smooth pursuit, we just chose the parameters that maximized the smoothness.
However, a vision scientist may find it interesting to experiment with other parameter choices when studying subtler eye motions.
Indeed, one can imagine a future system that dynamically adjusts the tuning parameters over time based on the motion of the eye as it is being tracked.
}

\changed{
\subsection{Towards a mobile, event-based tracker for AR/VR}
We built a head-mounted prototype of our eye tracking system to demonstrate its potential for miniaturization.
We designed a custom mount that sits on a user's head and holds a $22 \times 22$~mm DVS and lens connected to a battery-powered Raspberry Pi.
Figure~\ref{fig:sensor} shows our setup. It is slightly different from the one we used in our main experiments, since the sensor is lower resolution ($128 \times 128$ pixels) and does not output frames.
However, the DAVIS346 chip used in our desktop mounted experiments is already small enough ($25 \times 25$~mm) to fit on the same mount; this would require a custom PCB, which we will soon be able to demonstrate, along with recorded data from our setup and our custom 3D mount\footnote{\href{http://angelopoulos.ai/blog/posts/ebv-eye}{See project webpage: http://angelopoulos.ai/blog/posts/ebv-eye/}}.

Particularly given the relative ease of building a miniature event-based near-eye gaze tracker, we believe the technology could benefit the AR/VR community.
We discussed the speed advantage in Section~\ref{subsec:10khz}, but there are further benefits.
For example, AR/VR systems must conserve power to extend battery life, and event-based eye tracking consumes less power.
To see why, notice that during fixation and small eye motions, the sensor induces fewer events and incurs less processing, whereas each pixel in a frame must always be acquired and processed in a traditional system.
Furthermore, AR systems should perform well in a variety of indoor and outdoor lighting conditions; event-based cameras have a dynamic range of $\approx 130 dB$, much more robust than the traditional camera.
The future of eye-tracking undoubtedly involves mobile, untethered, always-on AR/VR glasses, so the efficiency of event-based sensors makes them a natural choice.
As a starting direction for interested future researchers, we note that the high update rate of our system would likely enable prediction of saccadic landing zones and durations, with AR/VR applications in efficient predictive foveated rendering~\cite{guenter2012foveated,konrad2019gaze} and autofocal AR~\cite{padmanaban2018autofocals,chakravarthula2018focusar}.
The most notable remaining challenge is slippage: head-mounted eye tracking systems often require 3D modeling of the user's head and face pose or recalibration to account for relative movement between the eye tracker and the head during use.
Our evaluations in Section~\ref{sec:results} do not account for slippage, which is often the primary driver of error in head-mounted eye tracking products.
However, given the longstanding work on event-driven 3D algorithms like SLAM~\cite{mueggler2017event}, we are optimistic that future researchers will user our platform as a starting point to build event-based slip compensation algorithms too.
As a parting conjecture, since the polynomial mapping between the pupil center and \changed{gaze point} is quite simple, an automatic calibration scheme may be possible by selecting an appropriate nearest-neighbor from a bank of users; the neighbor might be chosen based on the glint and ellipse parameters, for example.
}
%Such applications include AR/VR \cite{albert2017latency,patney2016towards,guenter2012foveated,padmanaban2017optimizing,konrad2019gaze,sitzmann2018saliency}, vision science experiments \cite{mele2012gaze}, and the recently demonstrated autofocal eyeglasses~\cite{padmanaban2018autofocals}.

%\textcolor{red}{NOTE: TALK ABOUT TINY EYE MOTIONS SUCH AS CORRECTIVE MICROSACCADES WHICH WE RELIABLY SEE IN OUR DATA.}

\acknowledgments{
A.N.A. was supported by a National Science Foundation (NSF) Fellowship and a Berkeley Fellowship. J.N.P.M. was supported by a Swiss National Foundation (SNF) Fellowship (P2EZP2 181817), G.W. was supported by an NSF CAREER Award (IIS 1553333), a Sloan Fellowship, by the KAUST Office of Sponsored Research through the Visual Computing Center CCF grant, and a PECASE by the ARL. Thanks to Stephen Boyd and Mert Pilanci for helpful conversations.}

\bibliographystyle{abbrv-doi}

\bibliography{bibliography}
\clearpage
\appendix
\section{Formulation of the problem for $\mathcal{P}$ and $\mathcal{C}$}
%%%%%%%%% BODY TEXT

%
\subsection{Updating and fitting the parabola $\mathcal{P}$}
The location of points on the parabola representing the eyelash in the image plane satisfy the quadric equation:
\begin{align}
E_{\mathcal{P}}(x,y) &= 0 \\
 \text{with}\quad E_\mathcal{P}(x,y) &=  a'\,y^2 + g'\,y + d' - x   \nonumber
\end{align}
Here, we assume the parabola can be written with $y$ as a function of $x$.
This assumption is valid as long as the eye is ``upright", that is, it is not excessively rotated in the image-space. The parabola is thus parameterized as $\mathcal{P}=(a',g',d')$ and it can be fitted similarly to the ellipses representing the pupils by solving:
\begin{equation}
    \mathcal{P}^* = \underset{\mathcal{P} \in \mathbb{R}^3}{\arg \min} \sum_{(x,y)\in\mathcal{D'}} E_{\mathcal{P}}(x,y)^2
\end{equation}
In which $\mathcal{D}'$ is the set of points belonging to the parabola, which is analogous to the set $\mathcal{D}$ for the ellipse representing the pupil. Again, the solution is simply $\mathcal{P}^* = A'^{-1}\,b'$ with
\begin{equation}
     A' = \sum_{(x,y)\in\mathcal{D}_{\mathrm{img.}}} v'^1_{x,y}\, v'^{1_{x,y}\; \intercal},\quad 
     b' = \sum_{(x,y)\in\mathcal{D}_{\mathrm{img.}}} v'^2_{x,y} \label{eq:AandB_img} 
\end{equation}
in which $v'^1_{x,y}$ is now:
\begin{equation}
 v'^1_{x,y} = (y^2, y, 1)
\end{equation}
and $v'^2_{x,y}$ is different (due to the asymmetry in x and y):
\begin{equation}
 v'^2_{x,y} = x\,v'^{1\; \intercal}_{x,y} = (x\,y^2, x\,y, x)\label{eq:v_img}
\end{equation}
Lastly, we need to describe how the points are selected in $\mathcal{D}'$. Similarly to $\mathcal{D}=\mathcal{D}_{\mathrm{evt.}}\cup\mathcal{D}_{\mathrm{img.}}$, we define $\mathcal{D}'=\mathcal{D}'_{\mathrm{evt.}}\cup\mathcal{D}'_{\mathrm{img.}}$. We detect the points belonging to the eyelash in an image by running a Harris corner detector on an image in which all gray values outside of $[t_1,t_2]$ have been clipped. Then all the candidate Harris corners further than a certain radius $\rho'$ from the currently estimated pupil center are discarded, and all the points in the lower half of the image are also discarded. This is:
\begin{equation}
\begin{aligned}
    \mathcal{D}'_{\mathrm{img.}} = \Big\{(x,y) \; | \;  & (x,y) \in\mathrm{HarrisCorner}\circ\mathrm{clip}(I,t_1,t_2), \\ 
    &\text{and } \lVert (x,y) - (x_e,y_e)\rVert^2 < \rho', \\
    &\text{and } y<\frac{\text{rows}}{2}\Big\}
\end{aligned}
\end{equation}
The events considered in the fit for the parabola are also those falling with a radius $\delta$ of the currently estimated parabola.
\begin{equation}
    \mathcal{D}'_{\mathrm{evt.}} = \Big\{(x,y) \;\lvert\; |E_{\mathcal{P}}\big((x,y)\big)| < \delta \Big\}
\end{equation}
\subsection{Updating and fitting the circle for the glint $\mathcal{C}$}
Regressing the glint is a subcase of regressing the ellipse in which one can choose the scaling $a=b=1$, and has $h=0$. The location of points on the circle representing the glint in the image plane satisfy the quadric equation:
\begin{align}
E_{\mathcal{C}}(x,y) &= 0 \\
 \text{with}\quad E_\mathcal{C}(x,y) &=  x^2 -2\,x\,c_x - 2\,y\,c_y + (c_x^2 + c_y^2 - r^2)  \nonumber
\end{align}
Hence we have to fit the parameters $\mathcal{C}=(c_x,c_y,r)$.
The points in an image to be selected to update the fit of the glint are those that exceed a certain threshold $t_3$ and that are less than a certain distance $\rho''$ from the current pupil center estimate.
\begin{equation}
\begin{aligned}
    \mathcal{D}''_{\mathrm{img.}} = \Big\{(x,y) \; \lvert \;  
        & (x,y) \in H_{t_3}(I(x,y)) \\
        &\text{and } \lVert (x,y) - (x_e,y_e)\rVert^2 < \rho'' \Big\} \\
\end{aligned}
\end{equation}
Events are those falling less than $\delta$ away from the currently estimated glint:
\begin{equation}
    \mathcal{D}''_{\mathrm{evt.}} = \big\{ (x,y) \;\lvert \;  | E_{\mathcal{C}}(x,y) | < \delta \big\}
\end{equation}

\section{Additional details about the fitting of $\theta$ for the regressor}
The regressor maps parameters of $\mathcal{M}$ to screen coordinates $x_s,y_s$.
%
%\subsection{Simple regressor model}
In its simplest form, also yielding the lowest computational load, our regressor is a second order polynomial that takes as input, the pupil center $(x_e,y_e)$ extracted from the parameters $\mathcal{E}$, and outputs the point $(x_s,y_s)$  in screen coordinates that the user is supposed to look at. Thus, our regressor is a multivariate vector-valued function:
\begin{equation}
    G_\theta(\mathcal{E}) = \begin{pmatrix}
                       x_s \\
                       y_s
                     \end{pmatrix}
     = \begin{pmatrix}
       G_{\theta^1}|_x(\mathcal{E})\\
       G_{\theta^2}|_y(\mathcal{E}) 
     \end{pmatrix} 
\end{equation}
In which the coordinate functions $G_{\theta^1}|_x(\mathcal{E})$ (on x) and $G_{\theta^2}|_y(\mathcal{E})$ (on y) are second order polynomials with parameters $\theta^1$ and $\theta^2$:
\begin{equation}
    G_{\theta^i}|_{x/y}(x_e,y_e) = \alpha_i\,x_e^2 + \gamma_i x_e\,y_e + \beta_i\,y_e^2 + \epsilon_i\,x_e + \zeta_i y_e + \eta_i 
\end{equation}
The parameters $\theta^1$ and $\theta^2$ are fitted solving the following linear least squares (G's are linear in their coefficients):
\begin{align}
    \text{arg min}_{\theta^1}\:\lVert  G_{\theta^1}|_{x}(x_e,y_e) - x_s \rVert^2 \\
    \text{arg min}_{\theta^2}\:\lVert  G_{\theta^2}|_{y}(x_e,y_e) - y_s \rVert^2
\end{align}
The regression is supervised by pairs $\{\big((x_e,y_e),(x_s,y_s)\big)\}$ produced during the calibration procedure in which $(x_s,y_s)$ points are presented to the user, and $(x_e,y_e)$ are obtained from the ellipse fit in image space:
\begin{align}
    x_e = \frac{2\,b\,g - h\,f}{h^2 - 4\,a\,b},\,\text{and } 
    y_e = \frac{2\,a\,f - h\,g}{h^2 - 4\,a\,b}
\end{align}
From the parameters $\mathcal{E}=(a,h,b,g,f,d)$
%
%\subsection{Full regressor model}
%
\section{Definitions used for accuracy and precision in our experiment}

In the main text, our ``accuracy" results are calculated using the ISO 5725 definition of ``trueness": the closeness of agreement between the arithmetic mean of a large number of test results and the true or accepted reference value, this is:
\begin{equation}
    \mathrm{Accuracy} = \frac{1}{n}\sum_{i=1}^{n}l_i
\end{equation}
where $n$ is the total number of estimates, and $l_i$ is the $L_2$ norm of the difference between our estimated gaze direction $\hat{d}_i=(\hat{\phi}_i,\hat{\theta}_i)$ and the true gaze angle $d_i=(\phi_i,\theta_i)$:
\begin{equation}
    l_i = \lVert \hat{d}_i - d_i \rVert_2
\end{equation}
Precision is defined by ISO 5725 as: ``the closeness of agreement between test results". We quantify this by computing the empirical standard deviation of the $\hat{d}_i$'s: 
\begin{equation}
    \mathrm{Precision} = \sqrt{\frac{1}{n-1}\sum_{i=1}^{n}(\hat{d_i}-\bar{\hat{d}})^2}
\end{equation}
With $\bar{\hat{d}}$ being the empirical mean of all $\hat{d}_i$.

\section{Calculating $\theta$ and $\phi$ from $x_s$ and $y_s$}

The relationship between the horizontal angle $\theta$, the vertical angle $\phi$, and the screen coordinates which a user looks at, ($x_s,y_s$), is trigonometric because the screen is at a fixed distance. Although the gaze vector seemingly gives information about 3D space, it is in fact only a two-dimensional quantity, because it has no (or unit) magnitude. With ($c_x,c_y$) being the center of the screen at a fixed distance $D$, The conversion is calculated simply as:
\begin{equation*}
\begin{split}
    \theta=\frac{180}{\pi}\tan^{-1}(|x_s-c_x|/D) \\
    \phi=\frac{180}{\pi}\tan^{-1}(|y_s-c_y|/D)
\end{split}
\end{equation*}

\section{Blink Detector}
We use a blink detector to classify which frames are likely to be blinks and remove them from our accuracy and precision calculations. The blink detector operates on a simple principle: during a blink, our fitted ellipse to the pupil will deform drastically in shape, often becoming long and thin to fit the dark line of the eyelashes. This blink detector uses changes in \textit{eccentricity}, or the ratio between the major and minor radius of the fitted ellipse, to identify blinks. However, raw changes in eccentricity can vary widely when the eye is looking at different areas on the screen; so, an adaptive threshold must be computed to identify relatively large changes in eccentricity. Specifically, the eccentricities $r_{1:n}$ of the last $n$ fitted ellipses are stored in a vector, $R_{n}$. The sample mean $\mu_n$ and sample standard deviation $\sigma_n$ of $R_{n}$ are calculated. Then, when a new frame comes in at time $n+1$, the eccentricity of the fitted ellipse from that frame, $r_{n+1}$, is compared to $\mu_n+\lambda\sigma_n$, with $\lambda>0$ being a tunable parameter. If $r_{n+1} > \mu_n+\lambda\sigma_n$, then that frame will be classified as a blink, and is not considered in accuracy and precision calculations. In addition, a small number $k$ of following frames are also classified as blinks, since a blink takes on average $3-4$ frames to complete. Then, $r_{2:n+1}$ is assigned to $R_{n+1}$ to preserve a constant buffer length, and the blink detector considers the next frame ($r_{n+2}$). Finally, we trim the data at the 2.5\% level to remove outliers. This blink detection method is advantageous in that it adaptively computes how large of an eccentricity change constitutes a blink based on statistics of our ellipse data; it is also conservative, in that if eccentricity changes in a certain region of gaze directions are high on average, they will not be classified as blinks.

\changed{
\section{Empirical smoothness}
Consider a sequence of gaze estimates $\left\{ \Big(x_s^{(t)}, y_s^{(t)}\Big)\right\}_{t=1}^{T}$ at evenly spaced times $t=1,...,T$.
We quantify the empirical smoothness of the eye as
\begin{equation*}
    \mathrm{smoothness} = \frac{1}{T-1}\sum_{t=2}^T \frac{1}{\Big\lvert\Big\lvert\left(x_s^{(t)},y_s^{(t)}\right) - \left(x_s^{(t-1)},y_s^{(t-1)}\right)\Big\rvert\Big\rvert_2}.
\end{equation*}

When the eye center does not change much between increments, the smoothness value is high.
The smoothness value is dependent on the time scale; we would expect that at our high update rate, the eye's motion should be smooth because the increments of time are small. 
However, we would not expect smooth motion at the 30Hz rate of frames.
Nonetheless, because we cannot know the true physiological smoothness of the eye's motion, optimizing parameters based on the empirical smoothness is an assumption.
We discuss the limitations of this assumption in Section 6 and suggest how future work may lift it.

\section{Frame only ablation}
\label{app:frameonly}
%===========================================
\begin{figure}[t!]
    \centering
    \hspace{-0.5cm}\includegraphics[width=0.5\textwidth]{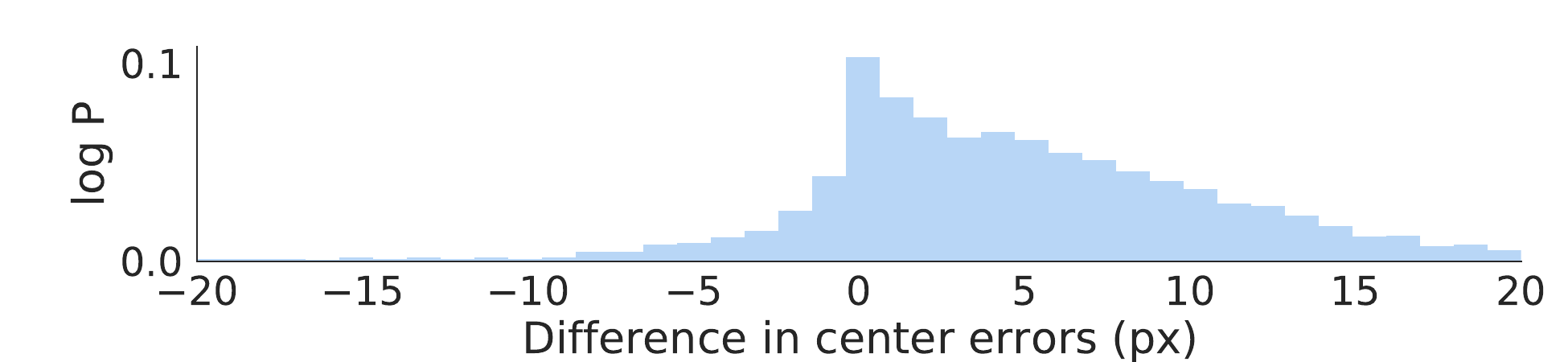}
    \caption{\textbf{The difference in pupil center error of the frame-only ablation} \changed{is plotted as a histogram.} See Appendix~\ref{app:frameonly} for details.}
    \label{fig:frameonly}
\end{figure}
%===========================================

We consider the relative error our system would incur if events were not used.
In particular, we define a simple {\em frame-only ablation}: rather than track the pupil between frames, we assume it does not change position between frames.
This frame-only ablation is equivalent to traditional (non-predictive) eye tracking.
For the majority of frames in our dataset, the eye is fixated, and in that case, there is not much difference between our system and the frame-only ablation.
Quantitatively, we confirm this: the mean IOUs of our system and the frame-only ablation are identical on such frames.
It is more informative to discuss the relative performance of the frame-only ablation on saccadic motions.
For every frame taken during a saccade, we calculate $d_{frame}$ as the error of the pupil center estimate of the frame-only ablation, and $d_{event}$ as the error of the pupil center estimate of the full system.
Then, in Figure~\ref{fig:frameonly}, we plot a histogram of $d_{frame}-d_{event}$.
Most of the histogram's mass is on positive numbers, illustrating the degradation in performance caused by the frame-only ablation.

Although we do not implement it here, one might also consider an event-only baseline.
An event-only ablation of our Algorithm 1 would need to be re-initialized after every blink, since the event-based estimate cannot recover from bad prior estimates of the pupil.
However, one can imagine an event-only baseline where events are stored and, for example, a Hough transform is performed to identify the pupil from the pooled event data only.
Various tricks could be performed to improve the performance and efficiency of such an algorithm; instead of storing a ring buffer of events, for example, they could be used in an online Hough-transform much like Algorithm 1.
Furthermore, the fact that the pupil and iris are concentric could increase the method's robustness (since there would be two nearby peaks in Hough space).
We hope these ideas might be useful to a future researcher working with an event-only sensor.
}

\end{document}